\documentclass[twoside,11pt]{article}
\usepackage[preprint]{jmlr2e}
\usepackage[T1]{fontenc}
\usepackage[utf8]{inputenc}
\usepackage{amsmath,mathtools}
\usepackage{lastpage}
\usepackage{newtxtext}
\usepackage{newtxmath}
\usepackage{newtxtt}

\pdfglyphtounicode{tfm:txexs/parenleftbig}{0028 FE01}
\pdfglyphtounicode{tfm:txexs/parenrightbig}{0029 FE01}
\pdfglyphtounicode{tfm:txexs/bracketleftbig}{005B FE01}
\pdfglyphtounicode{tfm:txexs/bracketrightbig}{005D FE01}
\pdfglyphtounicode{tfm:txexs/bardblex}{2225 FE01}
\pdfglyphtounicode{tfm:txexs/parenleftBig}{0028 FE02}
\pdfglyphtounicode{tfm:txexs/parenrightBig}{0029 FE02}
\pdfglyphtounicode{tfm:txexs/parenleftbigg}{0028 FE03}
\pdfglyphtounicode{tfm:txexs/parenrightbigg}{0029 FE03}
\pdfglyphtounicode{tfm:txexs/bracketleftbigg}{005B FE03}
\pdfglyphtounicode{tfm:txexs/bracketrightbigg}{005D FE03}
\pdfglyphtounicode{tfm:txexs/ceilingleftbigg}{2308 FE03}
\pdfglyphtounicode{tfm:txexs/ceilingrightbigg}{2309 FE03}
\pdfglyphtounicode{tfm:txexs/braceleftbigg}{007B FE03}
\pdfglyphtounicode{tfm:txexs/bracerightbigg}{007D FE03}
\pdfglyphtounicode{tfm:txexs/parenleftBigg}{0028 FE04}
\pdfglyphtounicode{tfm:txexs/parenrightBigg}{0029 FE04}
\pdfglyphtounicode{tfm:txexs/braceleftBigg}{007B FE04}
\pdfglyphtounicode{tfm:txexs/bracerightBigg}{007D FE04}
\pdfglyphtounicode{tfm:txexs/hatwide}{02C6 FE01}
\pdfglyphtounicode{tfm:txexs/hatwider}{02C6 FE02}
\pdfglyphtounicode{tfm:txexs/parenlefttpA}{239B}
\pdfglyphtounicode{tfm:txexs/parenrighttpA}{239E}
\pdfglyphtounicode{tfm:txexs/parenleftbtA}{239D}
\pdfglyphtounicode{tfm:txexs/parenrightbtA}{23A0}
\pdfglyphtounicode{tfm:txexs/parenleftexA}{239C}
\pdfglyphtounicode{tfm:txexs/parenrightexA}{239F}
\pdfglyphtounicode{tfm:txexs/summationtext}{2211 FE01}
\pdfglyphtounicode{tfm:txexs/producttext}{220F}

\usepackage{booktabs,longtable,array,microtype}
\usepackage{xurl}
\usepackage[section]{placeins}
\hypersetup{hidelinks,pdftitle={Label Geometry and Certified Kernel Subset Refitting},pdfauthor={Yuha Song}}
\ShortHeadings{Certified Kernel Subset Refitting}{Song}
\firstpageno{1}
\newcommand{\R}{\mathbb R}
\newcommand{\calE}{\mathcal E}
\newcommand{\op}{\mathrm{op}}
\newcommand{\err}{\mathcal R}
\newcommand{\floorerr}{\mathfrak r}
\newcommand{\tr}{\operatorname{tr}}
\newcommand{\rank}{\operatorname{rank}}
\newcommand{\diag}{\operatorname{diag}}

\newcommand{\one}{\mathbf 1}

\allowdisplaybreaks

\begin{document}
\title{Label Geometry and Certified Kernel Subset Refitting}
\author{\name Yuha Song \email yuha@gatech.edu\\
\addr School of Computer Science, College of Computing\\
Georgia Institute of Technology\\
Klaus Advanced Computing Building, 266 Ferst Drive\\
Atlanta, GA 30332-0765, USA}
% arXiv preprint: no journal editor, publication record, or arXiv ID is asserted.
\maketitle
\begin{abstract}%
Kernel ridge regression refitted from selected responses need not minimize worst-case error relative to the full-response fit.
A supplied ellipsoid specifies the response magnitudes allowed in different directions.
Worst-case tests on subsets of one or two responses characterize exactly when the original ridge rule is optimal for every subset and query admitted by the kernel, including singular kernels.
Entrywise bounds control refitting error under partial shape information, while an indistinguishability construction shows the limits of using local risk gaps without signs.
Established recovery and convex-design tools yield bounds on the best subset of a given size, verified for stored matrices.
A score model with bounded coefficients supplies the required response class before responses are read.
Experiments on frozen text embeddings repeatedly show smaller worst-case error alongside worse realized prediction fidelity.
On the 30 geometries in the primary input setting with 2,048 reference inputs, exploratory selection using geometry alone has certified worst-case error within 1.045 times the same-budget optimum.
These guarantees concern the prescribed response class and fixed queries; they do not establish general predictive superiority.
\end{abstract}

\section{Introduction}
Kernel ridge regression (KRR) from selected responses loses information and
can add decoding error. Compatibility asks when unchanged-ridge refitting
minimizes worst-case deviation from the full-response predictor over an
independently supplied ellipsoid; classical recovery solves unrestricted decoding.

Section~\ref{sec:recovery} characterizes compatibility for every subset and
admissible query through finite singleton/pair risk tests, including singular
kernels. With positive kernel diagonal, compatible shapes rescale
$K+\lambda I$ by one positive factor per connected component of the kernel's
nonzero-entry graph. Entrywise bounds transfer local excesses or supplied
shape intervals to all subsets; an indistinguishability construction limits
unsigned local information. Fixed-query pair tests can miss larger-set defects.

Section~\ref{sec:power-selection} specializes established column selection
and convex design to same-budget bounds. At fractional proposals
the variational bound is a Frank--Wolfe gap certificate; arbitrary-candidate
outward verification supplies the computational contract.
Sections~\ref{sec:certificate-comparison} and~\ref{sec:shape-robustness}
compare relaxation strength/cost and shape robustness.
Section~\ref{sec:application-contract} supplies a constrained-head
response-budget and partial-shape contract.

All six primary radius contrasts favor higher powers; no fidelity gain survives
multiplicity adjustment, and power four worsens fidelity on two corpora.
Later directional and spectral comparisons retain adverse outcomes.
Guarantees concern supplied classes and fixed queries; membership and
new-query performance require separate evidence.

\section{Related Work}\label{sec:related-work}
Classical optimal recovery \citep{micchelli1977,bakhvalov1971,foucart2022}
supplies the ellipsoidal decoder and minimax radius; KRR--Gaussian-process
posterior-mean equivalence \citep{kanagawa2018} connects the matrices.
Minimax regularization chooses a rule under prescribed models/noise
\citep{foucartliao2023,foucartliaoveldt2023}; compatibility here instead fixes
the ridge rule and characterizes its admissible response geometries.
The zero-noise case of \citet[Proposition~2]{gama2020} gives the same decoder
with invertible sampled covariance. Expected squared loss uses residual trace;
worst-case loss uses its largest eigenvalue, with an opposite-sign argument
establishing nonlinear minimaxity.

The radius is a spectral generalized column subset selection (GCSS) objective
(Equation~\ref{eq:gcss-equivalence}). \citet{farahat2013} give the Frobenius
greedy updates recovered at $p=1$; \citet{ordozgoiti2022} give target-aware
leverage selection; \citet{caixuxu2026} study spectral GCSS through interlacing
polynomials. Conditional-covariance weak submodularity
\citep{daskempe2018,elenberg2018,chamonribeiro2017}, spectral design/Schatten
objectives \citep{sagnol2013,dudley2022}, and score-sum optimum bounds
\citep[Proposition~8]{krause2008} are established. Here conditional correlation
controls singular-residual gains; $p>1$ uses a power proxy, not exact marginals.
Continuous sensor design and gap certificates \citep{joshiboyd2009,jaggi2013}
likewise precede the verified refitting specialization.

Conditional Gaussian specifications require symmetry/scale compatibility
\citep{besag1974,verhoef2018}; nullspaces can hide regularizer geometry
\citep{alberti2021}. The scalar pair contrasts identify the additional defect
of a prescribed ridge refit, including singular kernels. Nystr\"om methods
may access all responses \citep{bach2013,elalaoui2015}; this problem restricts
response access. Information-gain/D-optimal criteria
\citep{krause2008,chen2026} target different losses. Radius, reference fidelity
and task performance therefore require separate evaluation.

\section{Problem Setup and Available Information}\label{sec:setup}
A decoder maps retained labels or numerical targets to specified-query
predictions. For a fixed set and prescribed class, the recovery radius is
the smallest achievable worst-case deviation from the full-response reference;
an attaining decoder is minimax.

Write $[n]=\{1,\ldots,n\}$. Let $K\in\R^{n\times n}$ be symmetric positive semidefinite, $n\ge2$, and let $\lambda>0$. Set
\begin{equation}
 G=K+\lambda I_n,\qquad A_\lambda=K G^{-1}.
 \label{eq:reference}
\end{equation}
For $q\ge1$ kernel queries $Z$, put $C=K_{Z,[n]}\in\R^{q\times n}$ and
$A_Z=CG^{-1}$; the reference is $A_Zy$ for $y\in\R^n$. Training evaluation
uses $C=K$. Kernel realizability requires
\[
 C=CK^\dagger K,\qquad K_{ZZ}-CK^\dagger C^\top\succeq0,
\]
where $K_{ZZ}$ is the query Gram block and $K^\dagger$ the pseudoinverse.
Nullspace restriction and completing the square give these conditions.
Generic recovery allows arbitrary $A$.

For $S\subseteq[n]$, write $m=|S|$, $T=[n]\setminus S$. Let $E_S\in\R^{n\times m}$ insert coordinates into $[n]$ in increasing index order, so
$y_S=E_S^\top y$, $K_{SS}=E_S^\top K E_S$, and $K_{TS}=E_T^\top K E_S$.
Literal refitting ($\mu>0$) is
\begin{equation}
 B_{Z,S,\mu}=C E_S(K_{SS}+\mu I_m)^{-1}E_S^\top.
 \label{eq:literal}
\end{equation}
Write $B_{S,\mu}$ at training points. Unchanged ridge means $\mu=\lambda$,
without reweighting/relabeling. Empty retention predicts zero, full retention
is exact at $\mu=\lambda$, and empty blocks have norm zero. Inverse formulas use positive-definite
blocks and $1\le m<n$; endpoints are handled separately.

\subsection{What the Decoder May Know}
Selection uses geometry, queries and the supplied class, without labels.
Deterministic $g:\R^m\to\R^q$ reads only $y_S$; omitted-label encodings are
excluded. Section~\ref{sec:sequential-access} permits sequential choices
using already-read labels.

For a fixed $S$, the Euclidean-label error and its best-decoder counterpart are
\begin{align}
 \err_Z(S,\mu)&=\sup_{\|y\|_2\le1}\|A_Zy-B_{Z,S,\mu}y\|_2
                =\|A_Z-B_{Z,S,\mu}\|_{\op},\label{eq:error}\\
 \floorerr_Z(S)&=\inf_g\sup_{\|y\|_2\le1}\|A_Zy-g(E_S^\top y)\|_2.\label{eq:floor}
\end{align}
Subset minimization is outside the supremum. A label-chosen penalty/shape
generally gives a nonlinear decoder, so one realized fit's norm is not its
worst-case error; Proposition~\ref{prop:shape-stability} bounds a finite family
uniformly. Selection is deterministic given geometry and an independently fixed seed.

For $W\succ0$, define the centered ellipsoid
\begin{equation}
 \calE_W=\{W^{1/2}z:\|z\|_2\le1\}.
 \label{eq:ellipsoid}
\end{equation}
Square roots are symmetric positive semidefinite; $W$ specifies a deterministic
class, not a covariance. Put $W_{SS}=E_S^\top WE_S$. Subscripts $W$ indicate
$\calE_W$; $\floorerr_{A,W}(S)$ denotes the generic optimum and $Z$ is omitted
at training points. Norms $\|\cdot\|_{\op},\|\cdot\|_F$ are operator/Frobenius.
Matrices are finite-dimensional; for $M\succ0$,
$\kappa(M)=\lambda_{\max}(M)/\lambda_{\min}(M)$. 

\par\noindent\emph{What the compatible shape protects.}
The choice $W=G=K+\lambda I$ has a deterministic signal-and-perturbation
interpretation, including when $K$ is singular:
\begin{equation}
 y^\top G^{-1}y
 =\min_{\substack{f\in\operatorname{range}(K),\ e\in\R^n\\f+e=y}}
 \left(f^\top K^\dagger f+\frac{\|e\|_2^2}{\lambda}\right).
 \label{eq:compatible-signal-budget}
\end{equation}
The minimum is attained at $f=KG^{-1}y$, $e=\lambda G^{-1}y$.
Writing any other feasible signal as $f+u$, $u\in\operatorname{range}(K)$,
cancels the cross terms and adds
$u^\top K^\dagger u+\|u\|_2^2/\lambda\ge0$.
Thus $\calE_G$ protects kernel signals plus arbitrary label perturbations
under a shared quadratic budget. This standard decomposition offers a
possible modeling rationale, neither estimating covariance nor proving
membership of observed labels. Different protection requires a different supplied shape.
For example, $G+\rho hh^\top/(h^\top G^{-1}h)$, $\rho>0$, additionally protects
a declared contrast $h\ne0$: it allows $y=z+t h$ with
$z^\top G^{-1}z+t^2(h^\top G^{-1}h)/\rho\le1$.

\subsection{Regularization Convention}
In the kernel's reproducing kernel Hilbert space $\mathcal H$, shift
$\lambda$ corresponds to $\sum_i(y_i-f(x_i))^2+\lambda\|f\|_{\mathcal H}^2$.
Fixing average-loss coefficient $\alpha$ instead gives $\lambda=n\alpha$,
$\mu=m\alpha$. Experiments retain $\lambda=n\alpha$ after selection.
Deterministic classes imply no sampling-based population guarantee.
\section{Information Loss, Refitting Distortion, and Compatibility}\label{sec:recovery}
This section separates unavoidable information loss from distortion caused by refitting.
\subsection{Elementary Recovery Baselines}
\begin{lemma}[Coordinate-observation baseline]\label{lem:floor}
For any $A\in\R^{q\times n}$ and fixed $S$, the best worst-case error from observing only $y_S$ on $\|y\|_2\le1$ is $\|AE_T\|_{\op}$. The linear decoder $g(z)=AE_Sz$ attains it. For any linear retained-label decoder $L\in\R^{q\times m}$,
\begin{equation}
 (A-LE_S^\top)(A-LE_S^\top)^\top
 =AE_T E_T^\top A^\top+(AE_S-L)(AE_S-L)^\top.
 \label{eq:split}
\end{equation}
\end{lemma}
\begin{proof}
For $\|v\|_2\le1$, responses $y=\pm E_Tv$ have identical zero observations, so one incurs error
at least $\|AE_Tv\|_2$. Taking the supremum proves the lower bound, attained
by the stated residual $AE_Ty_T$. Empty and full retention give
$\|A\|_{\op}$ and zero. Partitioning the residual columns proves \eqref{eq:split}.
\end{proof}

\begin{proposition}[Additional distortion from ordinary refitting]\label{prop:distortion}
For $1\le |S|<n$, with $A=A_Z$ and $\mu=\lambda$, set
\[
 H_S=K_{TS}(K_{SS}+\lambda I_m)^{-1}.
\]
Then
\begin{equation}
 (A_Z-B_{Z,S,\lambda})(A_Z-B_{Z,S,\lambda})^\top
 =A_ZE_T(I_{|T|}+H_SH_S^\top)E_T^\top A_Z^\top,
 \label{eq:amplification}
\end{equation}
and
\begin{equation}
 \floorerr_Z(S)\le\err_Z(S,\lambda)
 \le\floorerr_Z(S)\sqrt{1+\|H_S\|_{\op}^2}.
 \label{eq:inflation}
\end{equation}
\end{proposition}
\begin{proof}
From $A_ZG=C$, the selected and omitted residual columns are
$-A_ZE_TH_S$ and $A_ZE_T$. Their product gives \eqref{eq:amplification};
Lemma~\ref{lem:floor} and
$\|I+H_SH_S^\top\|_{\op}=1+\|H_S\|_{\op}^2$ give \eqref{eq:inflation}.
\end{proof}
Changing the decoder leaves the information floor in \eqref{eq:split};
changing $S$ can change it. A small floor need not imply small refitting
error. These are operator-norm statements.

\begin{lemma}[Ellipsoidal recovery baseline]\label{lem:ellipsoid}
Let $W\succ0$ and $1\le |S|<n$. For any linear reference $A$, the decoder
\begin{equation}
 g_W(z)=L_Wz,\qquad L_W=AW E_S(E_S^\top W E_S)^{-1}
 \label{eq:ellipsoid-decoder}
\end{equation}
minimizes the worst-case Euclidean output error on $\calE_W$ among all deterministic decoders $g:\R^{|S|}\to\R^q$. The squared optimum is
\begin{equation}
 \left\|A\left(W-WE_S(E_S^\top WE_S)^{-1}E_S^\top W\right)A^\top\right\|_{\op}.
 \label{eq:ellipsoid-risk}
\end{equation}
For a scalar reference $A$, $L_W$ is the unique optimal \emph{linear} decoder.
\end{lemma}
\begin{proof}
Put $V=E_S^\top W^{1/2}$ and $\Pi=V^\top(VV^\top)^{-1}V$, the orthogonal
row-space projector. The stated decoder leaves $AW^{1/2}(I-\Pi)z$; the residual
operator's squared norm is \eqref{eq:ellipsoid-risk}. Restricting to $z\in\ker V$ gives
indistinguishable opposite responses and the matching lower bound for any decoder.

For any linear decoder $L\in\R^{q\times m}$, put $R_*=(A-L_WE_S^\top)W(A-L_WE_S^\top)^\top$.
The identity $(A-L_WE_S^\top)WE_S=0$ eliminates both cross terms, giving
\begin{equation}
 (A-LE_S^\top)W(A-LE_S^\top)^\top
 =R_*+(L-L_W)W_{SS}(L-L_W)^\top.
 \label{eq:weighted-distortion}
\end{equation}
For scalar $A$, $R_*=\floorerr_{A,W}(S)^2$ and the second term is positive
unless $L=L_W$, because $W_{SS}\succ0$. This proves the claimed uniqueness.
\end{proof}
Nonlinear uniqueness can fail: for $W=I_2$, $A=(0,1)$ and $S=\{1\}$,
zero is the unique minimax linear decoder, but $g(s)=s^2/2$ is minimax too.
For $s\in[-1,1]$, its conditional error is $\sqrt{1-s^2}+s^2/2\le1$,
with equality at $s=0$.

\par\noindent\emph{Equivalent column-selection problem.}
The same radius is a generalized column subset selection (GCSS) objective.
Let $\mathbf B=W^{1/2}$ and $\mathbf T=W^{1/2}A^\top$. Then
\begin{equation}
 \floorerr_{A,W}(S)
 =\|(I-\mathbf B_{:S}\mathbf B_{:S}^{\dagger})\mathbf T\|_{\op}.
 \label{eq:gcss-equivalence}
\end{equation}
The retained-column projector is
$\Pi_S=W^{1/2}E_SW_{SS}^{-1}E_S^\top W^{1/2}$, so the squared norm equals
\eqref{eq:ellipsoid-risk}. Use $\Pi_\varnothing=0$, $\Pi_{[n]}=I$ at endpoints.

The ellipsoidal recovery argument also extends columnwise. For $\ell\ge1$, let
$Y=W^{1/2}\Xi$, $\Xi\in\R^{n\times\ell}$, $\|\Xi\|_F\le1$.
Then $L_WY_S$ has minimax Frobenius error
$\|(A-L_WE_S^\top)W^{1/2}\|_{\op}$, the single-response radius.
The operator--Frobenius inequality proves the upper bound; restricting $\Xi$
to one nonzero column in $\ker(E_S^\top W^{1/2})$ gives the opposite-sign lower
bound against any matrix decoder. This joint budget differs from independent
column budgets; neither guarantees classification without a margin condition.

\subsubsection{Feasibility After Observing the Retained Labels}\label{sec:sequential-access}
For observed labels $s\in\R^m$ in Lemma~\ref{lem:ellipsoid}, put
$\nu=s^\top W_{SS}^{-1}s$. A completion in $\calE_W$ with $y_S=s$ exists
exactly when $\nu\le1$, with conditional minimax radius
\begin{equation}
 \inf_{a\in\R^q}\ \sup_{\substack{y\in\calE_W\\E_S^\top y=s}}
       \|Ay-a\|_2
 =\sqrt{1-\nu}\,\floorerr_{A,W}(S),
 \qquad a_*=L_Ws.
 \label{eq:conditional-radius}
\end{equation}
For $V=E_S^\top W^{1/2}$, completions decompose orthogonally as
$z=V^\top(VV^\top)^{-1}s+h$, $h\in\ker V$, with squared center norm $\nu$
and $\|h\|_2\le\sqrt{1-\nu}$. Their image center is $L_Ws$; opposite residuals
attain the stated radius. Completion is unique at $\nu=1$, absent at $\nu>1$
(no assigned radius). Feasibility certifies neither hidden membership nor
cross-validation (CV) shape estimates; observed-data candidates are unchanged.

For scalar output, put $r=\floorerr_{A,W}(S)$. Global minimaxity is equivalent to
\[
 |g(s)-L_Ws|\le r\{1-\sqrt{1-\nu}\}
 \qquad\text{for every }s\text{ with }\nu\le1.
\]
Conditional error is $r\sqrt{1-\nu}+|g(s)-L_Ws|$ and the zero slice forces
radius $r$. This includes $r=0$, $\nu=1$, requires $g(0)=0$, leaves infeasible
observations unrestricted, and covers the nonlinear example.

\par\noindent\emph{Deterministic sequential access.} 
Among deterministic rules choosing/stopping from geometry and already-read
labels, with at most $m$ distinct labels ($1\le m<n$), the optimal worst-case
error for fixed $A,W$ is $\min_{|S|=m}\floorerr_{A,W}(S)$.
The classical zero-transcript argument \citep{bakhvalov1971,micchelli1977}
reads $S_0$ on zero. Every admissible $y,-y$ with $y_{S_0}=0$ shares that
transcript and answer, forcing error at least $\|Ay\|_2$; its supremum is
$\floorerr_{A,W}(S_0)$, including empty $S_0$. Padding to $m$ cannot increase
it; a best fixed set with canonical decoding attains the stated minimum.
This global result uses no unread labels and gives no per-transcript or randomized expected-error bound.

\subsection{A Compatible Baseline}
\begin{corollary}[Ridge-compatible labels]\label{cor:compatible}
For every positive-semidefinite $K$, every query matrix $C$, and $W=K+\lambda I_n$, ordinary subset refitting with $\mu=\lambda$ is a minimax decoder on $\calE_W$ for each fixed subset $S$.
\end{corollary}
\begin{proof}
Since $A_ZW=C$ and $W_{SS}=K_{SS}+\lambda I_m$, Lemma~\ref{lem:ellipsoid} gives exactly $CE_S(K_{SS}+\lambda I_m)^{-1}$. Empty retention uses zero; full retention supplies the reference.
\end{proof}

\subsection{Local Tests for Universal Compatibility}
\label{sec:local-compatibility}
These tests characterize compatibility over every admissible query extension.
Auxiliary queries need not belong to the deployed batch or input domain;
worst-case excesses are oracle inputs, not training losses.
Section~\ref{sec:fixed-query-compatibility} distinguishes fixed-query pair tests.

For a nonempty retained set $S$ and scalar cross-vector $c$, write
\begin{align}
 z_S(c)&=c^\top G^{-1}WE_S-c_S^\top G_{SS}^{-1}W_{SS},
 \label{eq:pair-signed-residual}\\
 e_S(c)&=\sup_{y\in\calE_W}
 |c^\top G^{-1}y-c_S^\top G_{SS}^{-1}y_S|^2
 -\floorerr_{c^\top G^{-1},W}(S)^2
 =z_S(c)W_{SS}^{-1}z_S(c)^\top.
 \label{eq:pair-excess}
\end{align}
The last identity is the scalar case of \eqref{eq:weighted-distortion};
it also holds for full retention, with zero excess.
For $p=\{i,j\}$, $i<j$, define
\begin{equation}
 \sigma_{ij}=\begin{cases}1,&K_{ij}\ge0,\\-1,&K_{ij}<0,\end{cases}
 \quad c_p=K(e_i+\sigma_{ij}e_j)/\sqrt2,
 \quad \ell_p=G_{pp}^{-1}(c_p)_p.
 \label{eq:pair-query}
\end{equation}
These training-query combinations have coefficient norm one, belong to
$\operatorname{range}(K)$ and admit kernel extensions.

\begin{theorem}[Finite local characterization]\label{thm:compatibility}
Let $K\succeq0$, $\lambda>0$, $W\succ0$,
$\mathcal I=\{i:K_{ii}>0\}$, and $\mathcal J=[n]\setminus\mathcal I$.
The following are equivalent:
\begin{enumerate}
\item For every pair $p=\{i,j\}$ meeting $\mathcal I$,
 $e_{\{i\}}(c_p)=e_{\{j\}}(c_p)=e_p(c_p)=0$.
\item $W_{\mathcal I\mathcal J}=0$ and
 $W_{\mathcal I\mathcal I}=D_{\mathcal I}G_{\mathcal I\mathcal I}$,
 where $D_{\mathcal I}$ is positive diagonal and constant on each
 connected component of the nonzero-entry graph of $K$ on $\mathcal I$.
 The block $W_{\mathcal J\mathcal J}\succ0$ is unrestricted.
\item For every retained set and every admissible finite query matrix,
 literal refitting at $\lambda$ equals the canonical minimax decoder.
\end{enumerate}
Empty blocks are allowed. In particular, if every $K_{ii}>0$, the
compatible shapes are exactly $W=DG$ with componentwise constant $D$,
even when $K$ is singular.
\end{theorem}
\begin{proof}
Put $d_i=W_{ii}/G_{ii}$, $D=\diag(d_i)$, and $H=W-DG$, so $H_{ii}=0$.
Symmetry gives $W=GD+H^\top$. Canceling the common reference term in
the singleton and pair residuals yields
\begin{equation}
 z_{\{i\}}(c_p)-[z_p(c_p)]_i=\ell_{p,j}H_{ij},\qquad
 z_{\{j\}}(c_p)-[z_p(c_p)]_j=\ell_{p,i}H_{ji}.
 \label{eq:pair-contrast}
\end{equation}
Indeed, $W_{pp}=G_{pp}D_p+H_{pp}^\top$ and
$\ell_p^\top G_{pp}=(c_p)_p^\top$.
For $a=K_{ii}$, $b=|K_{ij}|$, $d=K_{jj}$, and
$\Delta_p=(a+\lambda)(d+\lambda)-b^2$,
\begin{equation}
 \ell_{p,i}=\frac{ad-b^2+\lambda(a+b)}{\sqrt2\Delta_p},\qquad
 \ell_{p,j}=\sigma_{ij}\frac{ad-b^2+\lambda(d+b)}{\sqrt2\Delta_p}.
 \label{eq:pair-sensitivity}
\end{equation}
Since $ad\ge b^2$, entry $i$ is nonzero exactly when $i\in\mathcal I$.
Zero excess forces each residual to zero. Thus \eqref{eq:pair-contrast}
sets both $H_{ij},H_{ji}$ to zero for active pairs. If only $j$ is active,
it sets $H_{ij}=0$; positive semidefiniteness implies $K_{ij}=0$, so
$H_{ji}=H_{ij}=W_{ij}=0$. Hence $H$ can remain only within
$\mathcal J\times\mathcal J$. On the active block, symmetry gives
$(d_i-d_j)K_{ij}=0$, proving condition 2.

Conversely, under condition 2 every active component has $W_C=d_CG_C$.
Substitution in \eqref{eq:ellipsoid-decoder} cancels $d_C$ on each block.
Admissible queries vanish on $\mathcal J$, and the zero cross-block prevents
inactive labels from supplying indirect information. This proves condition 3;
it implies condition 1 by \eqref{eq:pair-excess}. If $K=0$, every admissible
query is zero, and all three statements hold for arbitrary $W\succ0$.
\end{proof}

Universal minimaxity is equivalent to condition 3: scalar queries are included,
and Lemma~\ref{lem:ellipsoid} gives uniqueness among linear decoders.
For $K\succ0$, singleton tests over every scalar query already force
$WG^{-1}=\diag(W_{ii}/G_{ii})$. For singular $K$, queries span only
$\operatorname{range}(K)$ and pairs supply additional information.
When all $K_{ii}>0$, signed contrasts reconstruct the defect;
squared excesses yield the following bound.

\subsection{Compatibility for the Actual Query Operator}
\label{sec:fixed-query-compatibility}
At a deployed query matrix $C$, compatibility is weaker than the universal
condition. The weighted distortion identity directly gives the following test.

\begin{proposition}[Fixed-query compatibility and minimaxity]
\label{prop:fixed-query-compatibility}
Fix $C$, $A=CG^{-1}$, $W\succ0$, and $1\le |S|<n$. Put
\[
 L_0=C E_SG_{SS}^{-1},\qquad
 L_W=AWE_SW_{SS}^{-1},\qquad
 \Delta_S=AWE_S-L_0W_{SS},
\]
and let $R_S=A(W-WE_SW_{SS}^{-1}E_S^\top W)A^\top$,
$r_S^2=\lambda_{\max}(R_S)$.
Literal refitting equals the canonical decoder exactly when $\Delta_S=0$.
It is minimax at this fixed query operator exactly when
\begin{equation}
 \Delta_SW_{SS}^{-1}\Delta_S^\top
 \preceq r_S^2 I_q-R_S.
 \label{eq:fixed-query-minimax}
\end{equation}
For scalar output, these two conditions are equivalent. For vector output,
minimaxity can hold without equality of decoders.
\end{proposition}
\begin{proof}
The first assertion follows by multiplying $L_W-L_0$ by $W_{SS}$.
Equation~\eqref{eq:weighted-distortion} gives the literal error matrix
$R_S+\Delta_SW_{SS}^{-1}\Delta_S^\top$.
Its largest eigenvalue is at least $r_S^2$ and equals $r_S^2$ exactly under
\eqref{eq:fixed-query-minimax}. For $q=1$ its nonnegative added term must vanish;
for $q>1$, nonleading eigendirections can have slack.
\end{proof}

For $K=I_3$, $\lambda=1$, $C=(1,0,0)$, every
$W=\operatorname{diag}(1,B)$, $B\succ0$, gives canonical refitting for all sets.
Correlations within $B$ are invisible here but violate universal compatibility.

\par\noindent\emph{Fixed-query singleton and pair tests can miss a larger-set defect.}
Take $n=5$, $K=I_5+\one\one^\top$, $\lambda=1$, $C=e_1^\top$, and
\[
 F=\begin{pmatrix}
 0&1&-1&0\\1&0&0&-1\\-1&0&0&1\\0&-1&1&0
 \end{pmatrix},\qquad
 H=\operatorname{diag}(0,F),\qquad W=G+H/2.
\]
The query is realizable with $K_{ZZ}=5/6$, and $W$ has eigenvalues
$1,2,2,3,7$. Since $H\one=0$ and its first row vanishes, $AW=C$.
For any singleton or pair $S$, either $1\notin S$ and $C_S=0$, or
$1\in S$ and $W_{SS}=G_{SS}$. Thus all these fixed-query refits are canonical.
At $S=\{1,2,3\}$, however,
\[
 L_0=(2/5,-1/10,-1/10),\qquad
 L_W=(9/23,-2/23,-2/23).
\]
Their scalar squared-risk difference is $3/2300>0$; the canonical squared
radius is $6/161$, giving squared inflation $207/200$.
Universal tests use auxiliary queries and avoid this blind spot.
Equation~\eqref{eq:fixed-query-minimax} checks the actual set/operator with
supplied shape; deployed-query pair checks alone do not certify larger sets.

\subsection{From Local Excesses to Every Retained Set}
\label{sec:psd-pair-certificate}
\begin{theorem}[A quantitative local certificate]
\label{thm:quantitative-compatibility}\label{thm:psd-pair-certificate}
Suppose $K\succeq0$ and every $K_{ii}>0$, with no rank restriction,
$\lambda>0$, and $W\succ0$. Put $G=K+\lambda I$,
$d_i=W_{ii}/G_{ii}$, and $D=\diag(d_i)$.
Let nonnegative $\epsilon_{p,i},\epsilon_{p,j},\epsilon_p$ bound
$e_{\{i\}}(c_p),e_{\{j\}}(c_p),e_p(c_p)$, respectively. Define
\begin{equation}
 \eta^2=\sum_{i<j}\left\{
 \frac{W_{ii}(\sqrt{\epsilon_{p,i}}+\sqrt{\epsilon_p})^2}{\ell_{p,j}^2}
 +\frac{W_{jj}(\sqrt{\epsilon_{p,j}}+\sqrt{\epsilon_p})^2}{\ell_{p,i}^2}
 \right\}.
 \label{eq:pair-risk-certificate}
\end{equation}
On each graph component $C$, let $\bar d_C=|C|^{-1}\sum_{i\in C}d_i$,
and put $D_{\rm comp}=\diag(\bar d_{C(i)})$, $W_0=D_{\rm comp}G$.
Then $W_0\succ0$ is compatible and
\begin{equation}
 \|W-DG\|_F\le\eta,\qquad
 \|W-W_0\|_F\le B:=\eta+\|(D-D_{\rm comp})G\|_F.
 \label{eq:pair-shape-distance}
\end{equation}
If the weighted graph Laplacian with weights $K_{ij}^2$ has a positive
eigenvalue, let $\gamma_+$ be its smallest positive eigenvalue. Then
\begin{equation}
 B\le\left(1+\frac{\sqrt2\|G\|_{\op}}{\sqrt{\gamma_+}}\right)\eta.
 \label{eq:quantitative-compatibility}
\end{equation}
For an edgeless graph, $B=\eta$. If
$\zeta=B/\lambda_{\min}(W_0)<1$, then for every admissible finite query
matrix $C$ and every retained set $S$,
\begin{equation}
 \sup_{y\in\calE_W}\|A_Zy-CE_SG_{SS}^{-1}y_S\|_2
 \le \frac{\floorerr_{A_Z,W}(S)}{\sqrt{1-\zeta^2}}.
 \label{eq:pair-all-subset-transfer}
\end{equation}
The endpoint subsets use their usual rules. A label-independent
$\chi$-approximate minimizer of the recovery radius under $W_0$, for a fixed
query matrix and cardinality budget, has joint selection/refitting factor
$\chi\sqrt{(1+\zeta)/(1-\zeta)}$ relative to the best same-size subset
under $W$ for that query matrix.

The Frobenius aggregation can be strengthened. Set $U_{ii}=0$ and, for
$p=\{i,j\}$, set
$U_{ij}=\sqrt{W_{ii}}(\sqrt{\epsilon_{p,i}}+\sqrt{\epsilon_p})/|\ell_{p,j}|$
and define $U_{ji}$ analogously. Put
\[
 F=(D-D_{\rm comp})G,\qquad
 M_{ij}=\min\{U_{ij}+|F_{ij}|,\ U_{ji}+|F_{ji}|\}.
\]
For any positive diagonal $T$, choose
$0<\gamma_T\le\lambda_{\min}(T^{-1}W_0T^{-1})$ and
$\rho_T\ge\|T^{-1}MT^{-1}\|_{\op}$; the maximum row sum is an admissible
$\rho_T$. The relative conclusions hold whenever $\zeta_{\rm ent}<1$, with
\begin{equation}
 \zeta_{\rm ent}=\min\{\zeta,\rho_T/\gamma_T\}
 \quad\hbox{in place of }\zeta,\qquad
 \|W_0^{-1/2}(W-W_0)W_0^{-1/2}\|_{\op}\le\zeta_{\rm ent}.
 \label{eq:entrywise-local-certificate}
\end{equation}
One may minimize over several choices of $T$. For
$T=\diag(W_0)^{1/2}$, the computable choice
$\gamma_T=\lambda/\max_iG_{ii}$ is valid; the exact smallest eigenvalue of
the known normalized $W_0$ can sharpen it.
\end{theorem}
\begin{proof}
By \eqref{eq:pair-excess} and Cauchy--Schwarz,
$|z_{\{i\}}(c_p)|\le\sqrt{W_{ii}\epsilon_{p,i}}$ and
$|[z_p(c_p)]_i|\le\sqrt{W_{ii}\epsilon_p}$.
Equation~\eqref{eq:pair-contrast} and the triangle inequality give
\eqref{eq:pair-risk-certificate}'s bound on $\|H\|_F$.
The decomposition $W-W_0=H+(D-D_{\rm comp})G$ proves
\eqref{eq:pair-shape-distance}.
Since $(d_i-d_j)K_{ij}=H_{ji}-H_{ij}$,
\[
 \gamma_+\|d-\bar d_{C(\cdot)}\|_2^2
 \le\sum_{i<j}K_{ij}^2(d_i-d_j)^2
 \le2\|H\|_F^2.
\]
The componentwise graph Poincar\'e inequality thus proves
\eqref{eq:quantitative-compatibility}; isolated components contribute zero.
Finally, $(1-\zeta)W_0\preceq W\preceq(1+\zeta)W_0$.
Theorem~\ref{thm:compatibility} makes the literal decoder canonical for
$W_0$. Proposition~\ref{prop:shape-stability}, with comparison constants
$1/(1+\zeta)$ and $1/(1-\zeta)$, gives
\eqref{eq:pair-all-subset-transfer}. Equation~\eqref{eq:joint-selection-transfer}
gives the stated joint factor. These steps use existing shape-transfer principles.

For the refinement, the same contrasts give $|H_{ij}|\le U_{ij}$ although
$H=W-DG$ need not be symmetric. Since $E=W-W_0=H+F$ is symmetric,
$|E_{ij}|\le M_{ij}$. Thus, for every $x$,
$|x^\top Ex|\le |Tx|^\top T^{-1}MT^{-1}|Tx|
\le\rho_T\|Tx\|_2^2\le(\rho_T/\gamma_T)x^\top W_0x$.
This proves \eqref{eq:entrywise-local-certificate} and the same transfers.
For the stated diagonal normalization, componentwise scale cancellation gives
$T^{-1}W_0T^{-1}=\operatorname{corr}(G)\succeq
(\lambda/\max_iG_{ii})I$. The refinement applies elementary norm comparisons to local contrasts.
\end{proof}

\par\noindent\emph{A prescribed uncertainty-set interface.}
Instead of excess oracles, a model contract may specify the exact diagonal
$w_i>0$ and symmetric off-diagonal intervals
$W_{ij}\in[a_{ij},b_{ij}]$. Set
$U_{ij}=\max\{|a_{ij}-d_iG_{ij}|,|b_{ij}-d_iG_{ij}|\}$, $U_{ii}=0$, and
use $\eta=\|U\|_F$; the proof certifies every permitted positive-definite shape,
uniformly.
These are design tolerances, not unread-label estimates.
For example, let $K=I$, $\lambda=1$, $\diag(W)=2I$, and allow
$|W_{ij}|\le h$ only on disjoint matched pairs, fixing other off-diagonals
to zero. The normalized row bound gives $\zeta_{\rm ent}=h/2$ at any even $n$,
whereas the original aggregation gives $\zeta=h\sqrt n/2$ at the interval
extremes. For $h=1/5$, the former certifies inflation at most
$\sqrt{100/99}<1.00504$, including $n=100,1000,10000$; the latter fails its
strict gate at $n\ge100$. Training queries with one retained coordinate per
pair attain this factor when all pair correlations equal $h$.

\par\noindent\emph{An information limit for unsigned local excesses.}
For $n\ge4$ a power of two, let $H_n$ be the symmetric Sylvester Hadamard
matrix, $H_n^2=nI$, and put $a=h/(\sqrt n+1)$, $0<h<2$.
With $K=I$, $\lambda=1$, the two positive-definite shapes
\[
 W_{\rm s}=2I+a\{H_n-\diag(H_n)\},\qquad
 W_+=2I+a(\one\one^\top-I)
\]
have the same diagonal and exactly the same specified local excesses:
$e_p(c_p)=0$ and $e_{\{i\}}(c_p)=e_{\{j\}}(c_p)=a^2/16$ for every pair.
Yet $\|H_n-\diag(H_n)\|_{\op}\le\sqrt n+1$ bounds the all-query,
all-subset inflation for $W_{\rm s}$ by $(1-h^2/4)^{-1/2}$.
For $W_+$, training queries and $|S|=n/2$ give inflation
\[
 \frac{2-a+an/2}{\sqrt{(2-a)(2-a+an)}}
 \sim\sqrt{h/8}\,n^{1/4}.
\]
Indeed, with $b=2-a$ and $m=n/2$, literal squared radius is $(b+am)/4$,
while the Schur complement gives canonical squared radius
$b(b+an)/\{4(b+am)\}$. At $h=1/5$, $n=128$, the former shape has inflation
at most $1.00504$, whereas the latter attains approximately $1.06493$.
No valid rule using only $K$, the diagonal and these unsigned excesses can
approve 5\% inflation for the former while protecting all consistent shapes.
Signed information can separate them; tighter interval decisions must retain
the necessary sign restrictions.

The excess interface uses $K$, $\diag(W)$ and three upper excess bounds per
pair. With full $W$, direct comparison gives
\[
 \delta=\|W_0^{-1/2}(W-W_0)W_0^{-1/2}\|_{\op}
\le\zeta_{\rm ent}\le\zeta.
\]
For $\delta<1$, the no-weaker factor is $(1-\delta^2)^{-1/2}$; with a selected
set, Equation~\eqref{eq:fixed-query-minimax} directly checks the actual operator.
The local theorem characterizes structure and certifies restricted inputs;
it supplies no estimator of oracle excesses from retained labels.
It avoids $\lambda_{\min}(K)$ but retains local sensitivities and, when
inferring scales, graph conditioning. With
$K_t=\diag(1,t,t)$, $\lambda=1$, $0<t\le1$, and
$W=2I+h(e_2e_3^\top+e_3e_2^\top)$, $0<h<2$, every pair excess is zero.
The only nonzero singleton excesses are $h^2t^2/[4(1+t)^2]$ for pair
$\{2,3\}$. Yet the distance to the compatible diagonal class is $\sqrt2h$,
while $\|K_t\|_{\op}=1$, $\kappa(G_t)\le2$, and $W$ is fixed. The local
coefficients $t/[\sqrt2(1+t)]$ are therefore essential.
Graph-gap dependence is also necessary: for
$K_t=\left(\begin{smallmatrix}1&t\\t&1\end{smallmatrix}\right)$,
$0<t<1$, $\lambda=1$, $W=\diag(2,4)$, the defect is $\|W-DG_t\|_F=\sqrt5t$, but
\[
 \inf_{c>0}\|W-cG_t\|_F^2=\frac{8+20t^2}{4+t^2}\to2,
 \qquad \gamma_+=2t^2.
\]
With inactive coordinates, Theorem~\ref{thm:compatibility} remains exact,
but a quantitative relative guarantee also needs control of the unrestricted
$W_{\mathcal J\mathcal J}$ block. Its diagonal alone supplies no positive
lower spectral bound.

\par\noindent\emph{A singleton blind spot that pairs detect.}
\label{sec:training-query-checks}
For $K\succ0$, the weighted training-query singleton statistic obeys
\[
 T_{\rm train}:=\sum_{i,j}W_{ii}e_{\{i\}}(Ke_j)=\|QK\|_F^2,
 \qquad Q=WG^{-1}-D.
\]
Bounding $\|Q\|_F$ from $\sqrt{T_{\rm train}}$ incurs the factor
$1/\lambda_{\min}(K)$.
To see the distinction, take
$u=\one_4/2$, $K=uu^\top$, $\lambda=1$, and
\[
 F=\begin{pmatrix}0&1&-1&0\\1&0&0&-1\\-1&0&0&1\\0&-1&1&0\end{pmatrix},
 \qquad W=G+F/4.
\]
Here $W$ has eigenvalues $1/2,1,3/2,2$, $D=I$, and $FK=0$; all singleton
excesses vanish. The pair coefficients are $1/(3\sqrt2)$, while the
pair excess is $1/252$ when $F_{ij}=1$, $1/180$ when $F_{ij}=-1$, and
zero otherwise. Thus $\eta^2=6/7$, $B=\eta$, $W_0=G$, and the risk-only
original Frobenius certificate gives all-subset factor $\sqrt7$. Signed contrasts reconstruct
$F/4$ exactly; its relative operator norm is $1/2$, giving the sharper
factor $2/\sqrt3$. The risk-only bound above does not use those signs.
For $K_\epsilon=K+\epsilon I$ and $W_\epsilon=W+\epsilon I$, $\epsilon>0$,
\[
 T_{{\rm train},\epsilon}=\frac{\epsilon^2}{2(1+\epsilon)^2}\to0,
 \qquad \inf_{c>0}\|W_\epsilon-cG_\epsilon\|_F=1/\sqrt2,
\]
whereas the pair contrasts continue to recover the fixed defect $F/4$.

\section{Geometry-Aware Selection and a Spectral Lower Certificate}
\label{sec:power-selection}
For supplied $W\succ0$ and $A$, put
$V^S=W-WE_SW_{SS}^{-1}E_S^\top W$, $R_S=AV^SA^\top$, and
$\tau(S)=\lambda_{\max}(R_S)=\floorerr_{A,W}(S)^2$;
$R_\varnothing=AWA^\top$ and $R_{[n]}=0$.
For $U=[n]\setminus S$, let $Q_S=V^S_{UU}$,
$b_i=AV^Se_i$, and $d_i=V^S_{ii}$. At each step, for a fixed integer $p\ge1$,
retain a remaining coordinate maximizing
\begin{equation}
 s_i(S)=\frac{b_i^\top R_S^{p-1}b_i}{d_i}.
 \label{eq:power-score}
\end{equation}
Fix a tie rule; at $R_S=0$, stop and pad if needed. For $p=1$, $R_S^0$ is the
identity on $\operatorname{range}(R_\varnothing)$.
At $p=1$ this is \citet{farahat2013}'s trace-gain greedy; $p>1$ uses a
power-weighted gradient proxy, not necessarily maximizing the exact decrease
in $\tr(R_S^p)$. Larger powers emphasize large eigenvalues. Direct spectral
decrease has no positive general weak-submodularity ratio: for $W=A=I_2$,
either singleton leaves radius one, but both coordinates give zero.
For $W=G$, $A=CG^{-1}$, initialize $AV^\varnothing=C$ and
$R_\varnothing=CG^{-1}C^\top$; canonical decoding is literal refitting.
With $V=V^S$, $B=AV$, $b_i=Be_i$, update simultaneously from old quantities:
\[
 V\leftarrow V-V_{:i}V_{i:}/d_i,\qquad
 B\leftarrow B-b_iV_{i:}/d_i,\qquad
 R\leftarrow R-b_ib_i^\top/d_i.
\]
For $p>1$, $R\ne0$, $s>0$, $P=(R/s)^{p-1}$ preserves score rankings.
For comparator budget $k$, $0<\theta_S\le\gamma_S$ defined below, the lower
certificate is $[\tr(PR)-h_k(P)/\theta_S]_+/\tr(P)$: $h_k(P)$ sums the largest
$\min(k,|U|)$ values $b_i^\top P b_i/d_i$, with $R$ unscaled.
Here $[x]_+=\max\{x,0\}$; witness scaling cancels.
Storage and each update cost $O(n^2+qn+q^2)$ entries and operations.
Scores cost $O(qn)$ at $p=1$, $O(q^2n)$ at $p=2$; direct powers or
eigendecomposition add $O(pq^3)$ for larger fixed $p$.
Initial solves and optional certification are separate; no end-to-end training saving follows.

\begin{theorem}[Power-score selection and its lower certificate]
\label{thm:power-selection}
Let $W\succ0$, $A$ be fixed, $p\ge1$ an integer, and
$d=\operatorname{rank}(R_\varnothing)>0$. Define
\[
 \Gamma_S=\diag(Q_S)^{-1/2}Q_S\diag(Q_S)^{-1/2},\quad
 \gamma_S=\lambda_{\min}(\Gamma_S),\quad
 \gamma_0=\lambda_{\min}(\operatorname{corr}(W)).
\]
Here $\diag(M)$ keeps only $M$'s diagonal, and
$\operatorname{corr}(W)=\diag(W)^{-1/2}W\diag(W)^{-1/2}$.
For nonfull $S$, $\gamma_S\ge\gamma_0>0$. Write $v_p(S)=(\tr R_S^p)^{1/p}$ and,
for $0\le t\le n$, let $S_t$ be the subset after $t$ steps of \eqref{eq:power-score}.
For every comparator $T$ with $|T|\le k$, $1\le k\le n$,
\begin{equation}
 [v_p(S_t)-v_p(T)]_+
 \le e^{-\gamma_0t/(pk)}[v_p(\varnothing)-v_p(T)]_+.
 \label{eq:power-contraction}
\end{equation}
Consequently, with $\tau_k^*=\min_{|T|\le k}\tau(T)$,
put $c_k=\min\{d,n-k\}$ and $a_t=e^{-\gamma_0t/(pk)}$. Then
\begin{equation}
 \tau(S_t)\le (1-a_t)c_k^{1/p}\tau_k^*
                +a_t v_p(\varnothing).
 \label{eq:power-spectral-bicriteria}
\end{equation}

At any $S$ with $R_S\ne0$, let $h_k(S)$ be the sum of its largest
$\min(k,|U|)$ scores, and let $0<\theta_S\le\gamma_S$. Then
\begin{equation}
 L_{k,p}(S):=
 \frac{[\tr(R_S^p)-h_k(S)/\theta_S]_+}{\tr(R_S^{p-1})}
 \le\tau_k^*.
 \label{eq:power-lower-certificate}
\end{equation}
For $p=1$ the denominator is $d$ on the stated reduced target space;
the ambient output dimension also gives a valid weaker bound.
Set $L_{k,p}=0$ at zero residual or full retention. The maximum of
\eqref{eq:power-lower-certificate} over earlier steps remains a lower bound.
\end{theorem}
\begin{proof}
For disjoint $S,T$, Schur complementation gives
$G_T:=R_S-R_{S\cup T}=B_T(Q_S)_{TT}^{-1}B_T^\top$ with
$B_T=AV^SE_T$ and $G_i=b_ib_i^\top/d_i$. Normalization gives
$\sum_{i\in T}G_i\succeq\gamma_SG_T$.
To compare the correlations, first normalize $W$ to $Q=\operatorname{corr}(W)$.
Its Schur complement on $U$ is at least $\gamma_0 I$, by minimizing
$[z;x]^\top Q[z;x]$ over $z$. Its diagonal is at most one.
Diagonal normalization thus preserves the lower bound $\gamma_0 I$.

For $X\succeq Y\succeq0$ and integer $p\ge1$,
\begin{equation}
 \tr[X^{p-1}(X-Y)]\le\tr(X^p)-\tr(Y^p)
 \le p\tr[X^{p-1}(X-Y)].
 \label{eq:power-trace-sandwich}
\end{equation}
For $p\ge2$, diagonalize $Y$ and apply scalar Jensen to each eigenvector's
spectral measure of $X$ to obtain $\tr(YX^{p-1})\ge\tr(Y^p)$.
The upper bound follows from the supporting-line inequality for
$\tr(X^p)$; equivalently apply scalar Jensen in an eigenbasis of $X$.
For $p=1$ both are equalities. Neither commutation nor a positive smallest
target eigenvalue is required.

Put $P=R_S^{p-1}$, $v=v_p(S)$, and $v_T=v_p(T)$.
Since $R_{S\cup T}\preceq R_T$, H\"older's trace inequality gives
$\tr(PG_{T\setminus S})\ge v^{p-1}(v-v_T)$.
If $v>v_T$, the maximum score is at least
$\gamma_Sv^{p-1}(v-v_T)/k$. The first inequality in
\eqref{eq:power-trace-sandwich} bounds the actual power decrease below by
that score. Concavity of $x^{1/p}$ then gives
$v_p(S_{t+1})-v_T\le(1-\gamma_{S_t}/(pk))(v_p(S_t)-v_T)$.
If $v\le v_T$, monotonicity keeps the positive part zero.
Iteration proves \eqref{eq:power-contraction}; the stronger product
$\prod_{j<t}(1-\gamma_{S_j}/(pk))$ is also valid.
A spectral-optimal comparator can be padded to exactly $k$ coordinates
without changing its optimum. Since
$V^T=W^{1/2}(I-\Pi_T)W^{1/2}$ has rank $n-k$, its residual has rank at
most $c_k$. Also $v_p(T)\le v_p(\varnothing)$, so
\eqref{eq:power-contraction} implies
$v_p(S_t)\le a_t v_p(\varnothing)+(1-a_t)v_p(T)$.
Now $\tau(S_t)\le v_p(S_t)$ and
$v_p(T)\le c_k^{1/p}\tau_k^*$ prove
\eqref{eq:power-spectral-bicriteria}. At $k=n$ both the comparator
residual and $c_k$ are zero.

Finally, for every $|T|\le k$,
\[
 h_k(S)\ge\sum_{i\in T\setminus S}\tr(PG_i)
 \ge\theta_S\tr\{P(R_S-R_{S\cup T})\}
 \ge\theta_S\{\tr(R_S^p)-\tau(T)\tr(P)\}.
\]
Rearrange and minimize over $T$ to prove
\eqref{eq:power-lower-certificate}.
\end{proof}

\par\noindent\emph{Other witnesses and comparator overlap.}
For $R_S\ne0$, $1\le k<n$, nonzero $P\succeq0$, and
$0<\theta_S\le\gamma_S$, let $h_r$ sum the largest $r$ values
$b_i^\top P b_i/d_i$, with $h_0=0$. Put
$c_{k,S}=\min\{d,n-|S|,n-k\}$ and order the eigenvalues of $P$ decreasingly.
Then
\begin{equation}
 \frac{[\tr(PR_S)-h_{\min(k,|U|)}/\theta_S]_+}
      {\sum_{j=1}^{c_{k,S}}\lambda_j(P)}\le\tau_k^*.
 \label{eq:psd-witness-certificate}
\end{equation}
Indeed, a size-$k$ optimum has union residual $X=R_{S\cup T}$
with $0\preceq X\preceq\tau_k^*I$ and $\rank X\le c_{k,S}$.
In $P$'s eigenbasis, the diagonal of $X$ lies in $[0,\tau_k^*]$ and
sums to at most $c_{k,S}\tau_k^*$, giving
$\tr(PX)\le\tau_k^*\sum_{j\le c_{k,S}}\lambda_j(P)$.
Combine this with the score-sum inequality in the preceding proof.

More precisely, put $u=n-|S|$,
$\mathcal R=\{\max(0,k-|S|),\ldots,\min(k,u)\}$ and
$c_r=\min\{d,u-r\}$. The comparator has some $r=|T\setminus S|\in\mathcal R$;
its union residual has rank at most $c_r$ and added-score sum at most $h_r$.
Thus the same argument gives
\begin{equation}
 \min_{r\in\mathcal R}
 \frac{[\tr(PR_S)-h_r/\theta_S]_+}
      {\sum_{j=1}^{c_r}\lambda_j(P)}\le\tau_k^*,
 \label{eq:overlap-witness-certificate}
\end{equation}
with zero-dimensional terms defined as zero. Since $h_r\le h_{\min(k,u)}$
and $c_r\le c_{k,S}$, this never weakens the preceding fixed-set bound;
if $c_r=0$, the full-score inequality forces that bound to zero too.
Neither refinement need improve a maximum over earlier steps.
Ranks are exact: denominator-only truncation is invalid. Use a projected
witness consistently in trace, scores and denominator; replacing $d$ by $q$ is safe.
The standard rank-removal bound
$\tau_k^*\ge\lambda_{k+1}(R_\varnothing)$ follows because
$R_\varnothing-R_T\succeq0$ has rank at most $k$; eigenvalues beyond the
initial rank are zero. Maxima of these valid lower bounds remain valid.

\par\noindent\emph{Returned-set certificate.}
For $1\le k\le n$ and $0\le t\le n$, combine
\[
 \underline\tau_{k,t}
 =\max\left\{\lambda_{k+1}(R_\varnothing),
              \max_{0\le j\le t} L_{k,p}(S_j)\right\}
 \le\tau_k^*
\]
with \eqref{eq:psd-witness-certificate}, \eqref{eq:overlap-witness-certificate}
and, for $W=G$, $A=CG^{-1}$, Proposition~\ref{prop:krr-directional}.
Whenever $\underline\tau_{k,t}>0$, the returned set satisfies
\begin{equation}
 \frac{\floorerr_{A,W}(S_t)}
      {\min_{|T|\le k}\floorerr_{A,W}(T)}
 \le\sqrt{\frac{\tau(S_t)}{\underline\tau_{k,t}}}.
 \label{eq:returned-radius-certificate}
\end{equation}
Only $t=k$ compares equal budgets. Rank removal vanishes for $k\ge d$;
a zero combined lower bound gives no relative ratio, while $R_{S_t}=0$
independently certifies exact recovery. These bounds assume exact arithmetic and the prescribed shape.

The objective reduction $F_p(S)=\tr(R_\varnothing^p)-\tr(R_S^p)$ has
weak-submodularity ratio at least $\gamma_0/p$ by
\eqref{eq:power-trace-sandwich}. The $p=1$ trace principle and greedy/oversampling
arguments are established; supplied correlation geometry and singular residuals
remain explicit. At a common current set, exact trace-power marginal greedy decreases at least
as much as maximum-score selection, inheriting these guarantees.

For exact marginals, write $u=b_i/\sqrt{d_i}$, $a_j=u^\top R_S^j u$, and
$\Delta_p=\tr(R_S^p)-\tr[(R_S-uu^\top)^p]$.
Noncommuting trace expansion gives
\begin{align}
 \Delta_2&=2a_1-a_0^2,\label{eq:exact-p2-marginal}\\
 \Delta_4&=4a_3-4a_0a_2-2a_1^2+4a_0^2a_1-a_0^4.
 \label{eq:exact-p4-marginal}
\end{align}
At $p=2$, only $a_0$ is extra. At $p=4$, an $O(q^3)$ eigendecomposition and
$O(q^2n)$ shared basis transform give all moments by $O(qn)$ weighted sums.
The rules admit the stated dense asymptotic costs and guarantees;
Section~\ref{sec:exact-ablation} compares spectral error, realized fidelity and measured cost.

For $R_\varnothing\ne0$ and any $\epsilon>0$, choose
\[
 t=\min\left\{n,
   \left\lceil\frac{pk}{\gamma_0}
     \max\{0,\log(v_p(\varnothing)/\epsilon)\}\right\rceil\right\}.
\]
Before truncation, \eqref{eq:power-spectral-bicriteria} gives
$\tau(S_t)\le c_k^{1/p}\tau_k^*+\epsilon$; retention at $n$ is exact.
For $\tau_k^*>0$, $\eta>0$, necessarily $c_k\ge1$; choosing
$\epsilon=\eta c_k^{1/p}\tau_k^*$ gives radius factor
$\sqrt{1+\eta}\,c_k^{1/(2p)}$ against the $k$-label optimum.
This is bicriteria at $t>k$, with no saving at $t\ge n$.
Zero optimum permits only the additive bound; $R_\varnothing=0$ makes empty retention exact.
Use $\theta_S=\gamma_0$, or, for $W=K+\lambda I$,
$\theta_S=\lambda/\max_{i\notin S}d_i$ since Schur complementation gives
$Q_S\succeq\lambda I$. For $c_k\ge2$, integer $p=\lceil\log c_k\rceil$
gives $c_k^{1/p}\le e$ at budget factor $p/\gamma_0$; for $c_k=1$, $p=1$
gives dimension factor one.

\par\noindent\emph{An attained certificate.}
Let $n=dL$, $W=I_n$, $A=[I_d\ \cdots\ I_d]/\sqrt L$ with $L$ blocks.
At $S=\varnothing$, $\theta_\varnothing=\gamma_\varnothing=1$, $R=I_d$;
all scores are $1/L$ for every $p$. Every $k$-set leaves trace $d-k/L$,
hence largest eigenvalue at least $1-k/n$. If $d$ divides $k$, selecting
$k/d$ coordinates per row group equalizes residual eigenvalues, giving
$\tau_k^*=1-k/n=L_{k,p}(\varnothing)$.
Compatible KRR uses $K=\kappa I_n$, $\kappa>0$,
$C=(\kappa+\lambda)A$, and query Gram matrix $CK^{-1}C^\top$.
For $d\le k<n$ divisible by $d$, this positive exact bound improves zero rank removal.

\par\noindent\emph{Relation to spectral GCSS guarantees.}
In \eqref{eq:gcss-equivalence}, write
$W=U\diag(w_1,\ldots,w_n)U^\top$, $w_1\ge\cdots\ge w_n>0$, and let
$e_k$ denote the elementary symmetric polynomial. Set
$
 \beta=\frac{\max_j[U^\top\mathbf T\mathbf T^\top U]_{jj}}
                  {\|\mathbf T\|_{\op}^2},\quad
 \delta_k=\frac{e_k(w_1,\ldots,w_{n-1})}{e_k(w_1,\ldots,w_n)},\quad
 x_k=(1-\sqrt{\delta_k})^2.
$
For $\mathbf T\ne0$ and $1\le k<n$, \citet[Theorem~1.3]{caixuxu2026}
gives a $k$-set with squared radius at most
$\bigl(\sqrt{\beta x_k}+\sqrt{(1-\beta)(1-x_k)}\bigr)^2\tau(\varnothing)$
when $\delta_k\le(1-\sqrt\beta)^2$. This initial-residual bound and
Theorem~\ref{thm:power-selection}'s $\tau_k^*$/correlation bound imply no uniform
algorithmic superiority. On all 30 retained primary geometries ($W=G$,
$k/n=1/4$), $G$'s dominant eigenpair $(w,u)$ gives
$\beta\ge\|Cu\|_2^2/(w\tau(\varnothing))=0.99952$--$0.99988$,
violating $\beta\le0.0179492$ required by $\delta_k\ge3/4$.
The general largest-root guarantee is unaffected. The artifact derives both
explicit-bound orderings without GCSS execution.

\begin{proposition}[A compatible-KRR directional lower bound]
\label{prop:krr-directional}
For $W=G=K+\lambda I$, $A=CG^{-1}$ and $1\le k\le n$, fix a unit query direction $v$.
Put $c=C^\top v$, $r_v=c^\top G^{-1}c$, and let $h_{k,v}$ sum the
largest $k$ entries of $(c_i^2)_{i=1}^n$. If $c\ne0$, choose any
$a_v\ge c^\top K^\dagger c$ and define
\begin{equation}
 L_k^{\rm dir}(v)=\left[r_v-
 \frac{a_vh_{k,v}}{\lambda a_v+h_{k,v}}\right]_+
 \le\tau_k^*.
 \label{eq:krr-directional-lower}
\end{equation}
Set $L_k^{\rm dir}(v)=0$ when $c=0$ or $k=n$.
Kernel realizability ensures $c\in\operatorname{range}(K)$; in particular,
$a_v=v^\top K_{ZZ}v$ is admissible. The maximum over any chosen directions
can enter the returned-set certificate.
\end{proposition}

\begin{proof}
Positive-semidefinite Cauchy--Schwarz gives
$K\succeq cc^\top/(c^\top K^\dagger c)\succeq cc^\top/a_v$.
Inverse order and the rank-one inverse formula give, for $|T|\le k$,
$
 c_T^\top G_{TT}^{-1}c_T
 \le\frac{a_v\|c_T\|_2^2}{\lambda a_v+\|c_T\|_2^2}
 \le\frac{a_vh_{k,v}}{\lambda a_v+h_{k,v}}.
$
Now $\tau(T)\ge v^\top R_Tv=r_v-c_T^\top G_{TT}^{-1}c_T$.
The zero-cross-vector and full-retention cases give zero.
\end{proof}
This uses standard matrix inequalities and kernel realizability, without
dividing by a conditional-correlation lower bound.
For the realizable rank-one example $K=\one\one^\top$, $C=\one^\top$,
$K_{ZZ}=1$, the bound is attained at every $1\le k<n$:
$
 \tau_k^*=L_k^{\rm dir}(1)
 =\frac{n}{n+\lambda}-\frac{k}{k+\lambda}
 =\frac{\lambda(n-k)}{(n+\lambda)(k+\lambda)}>0.
$
Here $d=1$, so rank removal is zero.

\subsection{A Variational Certificate for Several Query Directions}
\label{sec:split-certificate}
Specializing established design relaxations \citep{joshiboyd2009} extends the
rank-one-exact bound to several directions without a conditional-correlation divisor.

\begin{proposition}[A variational same-budget witness]\label{prop:split-certificate}
Let $G\succ0$, $C\in\R^{q\times n}$, $W=G$, $A=CG^{-1}$, and $1\le k<n$.
Write $R_S=CG^{-1}C^\top-C_SG_{SS}^{-1}C_S^\top$. Choose $\beta>0$
with $K_\beta=G-\beta I\succeq0$, any nonzero $F\in\R^{q\times r}$,
and any $X\in\R^{n\times r}$. Put $Z=C^\top F$, $U=Z-K_\beta X$, and
let $h_k(U)$ sum the $k$ largest squared row norms of $U$. Then
\begin{equation}
 L^{\rm var}_k(F,X)=
 \frac{[\tr(Z^\top G^{-1}Z)-\tr(X^\top K_\beta X)
                  -h_k(U)/\beta]_+}{\|F\|_F^2}
 \le\min_{|S|\le k}\lambda_{\max}
       (CG^{-1}C^\top-C_SG_{SS}^{-1}C_S^\top).
 \label{eq:split-certificate}
\end{equation}
Witnesses may depend on geometry but are common to comparator subsets;
their maximum remains valid. The split $\beta$ changes neither geometry,
shape, retained sets nor predictors.
\end{proposition}
\begin{proof}
For $B_S=[E_S^\top K_\beta^{1/2},\sqrt\beta I]$,
$B_SB_S^\top=G_{SS}$. The stacked matrix with blocks
$K_\beta^{1/2}X$ and $U_S/\sqrt\beta$ maps to $Z_S$ under $B_S$.
Bounding the minimum preimage norm gives
\[
 \tr(Z_S^\top G_{SS}^{-1}Z_S)
 \le\tr(X^\top K_\beta X)+\|U_S\|_F^2/\beta
 \le\tr(X^\top K_\beta X)+h_k(U)/\beta.
\]
Subtract from $\tr(Z^\top G^{-1}Z)$ and use
$\tr(F^\top R_SF)\le\lambda_{\max}(R_S)\|F\|_F^2$.
Positive semidefiniteness permits clipping. The empty subset is immediate;
minimization proves the result even for singular $K_\beta$.
\end{proof}

For $z\in[0,1]^n$, $\sum_i z_i=k$, put $D_z=\diag(\sqrt{z_i})$, $D=D_z^2$.
The proposal $X=D_z(\beta I+D_zK_\beta D_z)^{-1}D_zZ$ satisfies
$\beta X=DU$, including zero weights. It minimizes the convex quadratic
over $H\in\R^{n\times r}$:
\begin{equation*}
 J_z(H)=\tr(H^\top K_\beta H)
       +\beta^{-1}\|D_z(Z-K_\beta H)\|_F^2,
 \qquad \min_H J_z(H)=\tr(Z^\top X(z)).
\end{equation*}
Indeed, $\beta X=DU$ zeros its $H$-gradient; substitution gives the minimum.
Since $J_z(H)$ is affine in $z$, $f(z)=\tr(Z^\top G^{-1}Z)-\min_H J_z(H)$
is convex. Differentiating $\beta X=D(Z-K_\beta X)$ gives
$\partial_i f=-\|U_{i:}\|_2^2/\beta$, including zero weights and singular
$K_\beta$, for conditional-gradient search.
Any $X$ is valid without fractional feasibility or convergence; nonorthogonal
$F$ uses its enclosed actual norm. Sections~\ref{sec:certificate-comparison}
and~\ref{sec:multidirectional-study} give the Frank--Wolfe/full-relaxation
comparison and eight-selector certification.

\par\noindent\emph{Transfer to refitting under a nearby shape.}
For power selection under compatible $W_0$ with
$(1-\zeta)W_0\preceq W\preceq(1+\zeta)W_0$, $0\le\zeta<1$,
ellipsoid inclusion bounds literal refitting's squared worst-case error by
\[
 \frac{1+\zeta}{1-\zeta}(1-a_t)c_k^{1/p}\tau_{k,W}^*
 +(1+\zeta)a_t v_{p,W_0}(\varnothing),
\]
Here $d$, $\gamma_0$, $c_k$, and $a_t$ refer to $W_0$;
no estimation of $W$ from retained labels is used.

\section{Robustness to Prescribed Shape Misspecification}\label{sec:shape-robustness}
The classical Kantorovich constant \citep{newman1960} sharpens elementary norm
comparison in this recovery setting; the inequality is not new and does not
certify a CV-estimated shape.

\begin{proposition}[Sharp fixed-subset sensitivity]\label{prop:shape-stability}
Fix $A\in\R^{q\times n}$, $1\le |S|<n$, and $W,\widehat W\succ0$ with
$a_0W\preceq\widehat W\preceq b_0W$, where $0<a_0\le b_0$.
Let $L_{\widehat W}=A\widehat W E_S(\widehat W_{SS})^{-1}$. Then
\begin{equation}
 \sup_{y\in\calE_W}\|Ay-L_{\widehat W}y_S\|_2
 \le \frac{a_0+b_0}{2\sqrt{a_0b_0}}\,\floorerr_{A,W}(S).
 \label{eq:shape-stability}
\end{equation}
The coefficient is optimal as a universal fixed-subset bound, including for
KRR training references. If a finite family of shapes obeys the same comparison,
any rule choosing among them using $y_S$ alone obeys the same bound.
\end{proposition}
\begin{proof}
Put $V=E_S^\top W^{1/2}$, $F=AW^{1/2}$,
$M=W^{-1/2}\widehat W W^{-1/2}$, and
$N=I-MV^\top(VMV^\top)^{-1}V$. The residual is $FN$ and
$a_0I\preceq M\preceq b_0I$. For the orthogonal projector $\Pi_0$ onto
$\ker V$, $VN=0$ gives $N=\Pi_0N$, while Lemma~\ref{lem:ellipsoid} gives
$\|F\Pi_0\|_{\op}=\floorerr_{A,W}(S)$.
In orthonormal coordinates adapted to $\operatorname{range}(V^\top)$ and
$\ker V$, the blocks of $M$ give
\[
 N=\begin{pmatrix}0&0\\-M_{21}M_{11}^{-1}&I\end{pmatrix},\qquad
 \|N\|_{\op}^2=1+\|M_{21}M_{11}^{-1}\|_{\op}^2.
\]
For a unit vector $x$ in the first block, put $h=x^\top M_{11}x$.
The inequality $M^2\preceq(a_0+b_0)M-a_0b_0I$ implies
$\|M_{21}x\|_2^2\le(b_0-h)(h-a_0)$, while
$\|M_{11}x\|_2\ge h$. Because $M_{11}\succ0$, the change of variable
$z=M_{11}x$ gives
$\|M_{21}M_{11}^{-1}\|_{\op}=\sup_{x\ne0}\|M_{21}x\|_2/\|M_{11}x\|_2$. Hence
\[
 \|M_{21}M_{11}^{-1}\|_{\op}^2
 \le\max_{a_0\le h\le b_0}\frac{(b_0-h)(h-a_0)}{h^2}
 =\frac{(b_0-a_0)^2}{4a_0b_0}.
\]
Completing the square gives this maximum at $h=2a_0b_0/(a_0+b_0)$;
$\|FN\|_{\op}\le\|F\Pi_0\|_{\op}\|N\|_{\op}$ proves the bound.
For sharpness take $K=I_2$, $\lambda=1$, $A=I_2/2$, $W=I_2$, $S=\{1\}$,
and
\[
 \widehat W=\frac1{a_0+b_0}
 \begin{pmatrix}
  2a_0b_0&\sqrt{a_0b_0}(b_0-a_0)\\
  \sqrt{a_0b_0}(b_0-a_0)&a_0^2+b_0^2
 \end{pmatrix}.
\]
Its eigenvalues are $a_0,b_0$; the $W$-radius is $1/2$ and the mismatched error is
$(a_0+b_0)/(4\sqrt{a_0b_0})$. The uniform bound holds pointwise for each
admissible $y$, so any $y_S$-based candidate choice preserves it.
\end{proof}
For $\kappa=b_0/a_0$, the factor $(\sqrt\kappa+1/\sqrt\kappa)/2\le\sqrt\kappa$.
Tight constants are the extreme eigenvalues of $W^{-1/2}\widehat W W^{-1/2}$;
proportional shapes give factor one, while loose constants may not. Also
\eqref{eq:weighted-distortion} gives $r^2\le e_W(L)^2\le r^2+
\|(L-L_W)W_{SS}^{1/2}\|_{\op}^2$, where $r=\floorerr_{A,W}(S)$ and
$e_W(L)=\|(A-LE_S^\top)W^{1/2}\|_{\op}$. This remains valid at zero radius;
vector outputs need not attain the upper bound.

\par\noindent\emph{After observing retained values.} For any output dimension, put
$r=\floorerr_{A,W}(S)$ and $\nu=s^\top W_{SS}^{-1}s\le1$. Then
\[
 \sup_{y\in\calE_W:\,y_S=s}\|Ay-L_{\widehat W}s\|_2
 \;\le\; r\left(\sqrt{1-\nu}+\frac{b_0-a_0}{2\sqrt{a_0b_0}}\sqrt{\nu}\right).
\]
In the proof's blocks, $F=(F_1,F_0)$, $\|F_0\|_{\op}=r$, and observations fix
$z_1$ with $\|z_1\|_2^2=\nu$. The conditional image has center $L_Ws$, radius
$r\sqrt{1-\nu}$ and displacement
$(L_{\widehat W}-L_W)s=F_0M_{21}M_{11}^{-1}z_1$.
The block bound and triangle inequality prove the result, including $r=0$,
$\nu=1$. Joint matrix responses use Frobenius norm and
$\nu=\tr(Y_S^\top W_{SS}^{-1}Y_S)\le1$.
For scalar output, conditional error equals
$r\sqrt{1-\nu}+|(L_{\widehat W}-L_W)s|$. Equality throughout is attained by
the sharpness example's second output: $A=(0,1/2)$, $r=1/2$, $L_Ws=0$,
$L_{\widehat W}s=(b_0-a_0)s/(4\sqrt{a_0b_0})$, $\nu=s^2$.
If $b_0>a_0$, its conditional-radius ratio diverges as $|s|\uparrow1$,
where the optimum becomes zero. Whole-ellipsoid comparison therefore gives
no uniform conditional ratio or unknown-membership certificate.

\subsection{Joint Selection and Decoding under Shape Comparison}\label{sec:joint-transfer}
Fix $1\le m<n$, reference $A$, and label-independent $W,\widehat W\succ0$ with
$a_0W\preceq\widehat W\preceq b_0W$, $0<a_0\le b_0$. Let a label-independent
subset $\widehat S$ of size $m$ satisfy
\[
 \floorerr_{A,\widehat W}(\widehat S)
 \le\chi\min_{|S|=m}\floorerr_{A,\widehat W}(S),\qquad \chi\ge1.
\]
Decode the retained labels using
$L_{\widehat W,\widehat S}=A\widehat W E_{\widehat S}
(\widehat W_{\widehat S\widehat S})^{-1}$. Then
\begin{equation}
 \sup_{y\in\calE_W}\|Ay-L_{\widehat W,\widehat S}y_{\widehat S}\|_2
 \le \chi\sqrt{b_0/a_0}\,
       \min_{|S|=m}\floorerr_{A,W}(S).
 \label{eq:joint-selection-transfer}
\end{equation}
Norm comparison bounds the left side by
$a_0^{-1/2}\floorerr_{A,\widehat W}(\widehat S)$. Apply the $\chi$-approximation
and $\floorerr_{A,\widehat W}(S)\le\sqrt{b_0}\floorerr_{A,W}(S)$ for each $S$,
by evaluating the $W$-optimal decoder on $\calE_{\widehat W}$.
Minimizing proves \eqref{eq:joint-selection-transfer}, including when the optimum is zero.
No flat spectrum is required, but approximation must target the same recovery
radius: log-det supplies no $\chi$, nor does CV certify $a_0,b_0$.

The factor $\sqrt{b_0/a_0}$ is sharp even at $\chi=1$: take $\kappa>1$,
$0<\delta<1-\kappa^{-1/2}$, $t=\sqrt\kappa(1-\delta)>1$,
$W=I_2$, $\widehat W=\diag(1,\kappa)$, $m=1$, $\lambda=1$, and
\[
 K=\diag\!\left(1,\frac1{2t-1}\right),\qquad
 A=K(K+I_2)^{-1}=\diag\!\left(\frac12,\frac1{2t}\right).
\]
Under $\widehat W$, index 2 has radius $1/2$, strictly below index 1's
$\sqrt\kappa/(2t)=1/[2(1-\delta)]$. Under $W$, index 1 is optimal at $1/(2t)$;
index 2 gives $1/2$. Both shapes use the same fixed-subset zero-fill decoder;
only selection changes. The limit $t\to\sqrt\kappa$ as $\delta\downarrow0$
separates the two sharp factors.

\section{Consistency Checks and a Prescribed-Shape Use Case}\label{sec:numerical}
Archived structural checks cover 38 kernels, 522 pair queries, 1,574 scalar
refits, 3,156 kernel/subset comparisons and 150 same-budget comparisons.
Separate direct-Schur checks use 27 small geometries and $p=1,2,3,4,8$:
all 4,860 original power-bound/prefix comparisons agree with exhaustive
optima within the declared tolerance. The largest scaled update discrepancy
is $1.04\times10^{-12}$. These checks predate the later rank refinements.
Proposition~\ref{prop:krr-directional} additionally passes 966 comparisons
over 13,056 subsets and 231 substitutions in its attained rank-one family.

The entrywise refinement passes 2,640 small-subset and 60 joint-selection
comparisons, exact rational contrast/positivity checks, the replicated sparse
family, and retained adverse dense cases. The information-limit construction
has 126 exact rational local-risk checks and 536 small all-subset comparisons.
An independent Schur replay reproduces all 73,692 shape-study rows below
within $8.66\times10^{-15}$. The variational witness has exact small-subset
checks and an independent adversarial audit covering dependent directions,
subnormals, cancellation, full retention and required input rejections.
All inputs, tolerances, failures and source dates remain recorded. These
finite checks support implementation consistency; the stated proofs establish
the mathematical claims, and observed matrix bounds additionally require
Appendix~\ref{app:verified-arithmetic}'s enclosure model.

\subsection{A Prescribed Contrast on Observed Inputs}
\label{sec:observed-shape-use}
A post hoc diagnostic takes the first twelve reference/query indices from
each corpus's original large-model, seed-720001 $n=512$ geometry. Preprocessing,
bandwidth and $\lambda=0.512$ stay fixed, defining a twelve-response reference.
For the most dissimilar reference pair, set $h=e_i-e_j$ and prescribe
\[
 W_\rho=G+\rho\,\frac{hh^\top}{h^\top G^{-1}h},
 \qquad \rho\in\{0,0.001,0.01,0.1,1,4\}.
\]
This label-independent contrast is prescribed, not estimated. A dated protocol
precedes the risk calculations. All 73,692 corpus/shape/subset comparisons use
local-excess formulas and direct numerical risks with fully known $W_\rho$.

\begin{table}[t]
\centering\small
\setlength{\tabcolsep}{4pt}
\begin{tabular}{rrrrr}
\toprule
$\rho$ & Frobenius bound & Refined bound & Largest ratio & Direct bound\\
\midrule
0 & 1.000000 & 1.000000 & 1.000000 & 1.000000 \\
0.001 & 1.000031 & 1.000007 & 1.000000 & 1.000000 \\
0.01 & 1.003077 & 1.000679 & 1.000012 & 1.000012 \\
0.1 & 1.593073 & 1.073912 & 1.001134 & 1.001136 \\
1 & --- & --- & 1.060569 & 1.060660 \\
4 & --- & --- & 1.341181 & 1.341641 \\
\bottomrule
\end{tabular}
\caption{Prescribed-shape diagnostic on three observed $n=q=12$
subgeometries. The ratio is literal error divided by canonical radius,
maximized over all 4,094 nontrivial subsets and then the three corpora.
The two local columns give the largest original and entrywise-refined
inflation bounds across corpora; --- means every corpus fails the strict
comparison gate. Direct comparison has the same
analytic factor in all three. Values are numerical diagnostics, not
outward-rounded certificates; ranges are not confidence intervals.}
\label{tab:shape-use-case}
\end{table}

Table~\ref{tab:shape-use-case} separates certification from actual distortion.
With a predeclared 5\% inflation tolerance, local and direct bounds
approve literal refitting at the evaluated $\rho\le0.01$. At $\rho=0.1$,
the entrywise refinement newly approves DBpedia14 alone; its largest factor
across corpora is 1.073912, down from 1.593073. Direct comparison approves
all three, while actual inflation remains below 0.114\%.
At $\rho=1,4$, the largest observed-subproblem inflation is 6.057\% and
34.118\%, while the local condition fails. The exhaustively enumerated $k=3$ optimal subset
changes only for AG News at these two values. At $\rho=4$, retaining its
baseline-optimal set and literal decoder costs a factor 1.30386 relative
to the new canonical optimum; changing only the decoder reduces this to
1.12461. Shape changes can therefore affect refitting and selection. Direct comparison
is preferable with full shape knowledge; the interval interface instead uses
entry bounds and an exact diagonal. Neither estimates unknown membership.

\subsection{A Constrained-Score Design Contract}
\label{sec:application-contract}
Fix $b>0$ without reading responses. For $Y=G^{1/2}\Xi$, $\|\Xi\|_F\le b$,
$|S|\le k$, put $\widehat Y^{(S)}=C_SG_{SS}^{-1}Y_S$.
A squared-radius upper $\overline\tau_S$ certifies uniform Frobenius error
at most $\varepsilon\ge0$ when $b\sqrt{\overline\tau_S}\le\varepsilon$.
A same-budget lower $L_k$ rules out every $k$-set when
$b\sqrt{L_k}>\varepsilon$; otherwise unresolved.
Known $\|Y_{i:}\|_2\le M$ and verified $G\succeq\mu I$, $\mu>0$, give
$b=M\sqrt{n/\mu}$ without omitted labels; $M=1$ covers all one-hot labelings.
These fixed-operator class guarantees imply neither hidden membership,
classification, acquisition cost nor one-hot impossibility.

For verified stored $K_\beta=G-\beta I=\Phi\Phi^\top$, let calibration $B_\rho$
have unit rows, orthogonal across disjoint pairs, with within-pair inner products
$|\rho_p|\le r<1$. Pair consecutive references without labels or semantic claims.
Put $M_\rho=[\Phi,\sqrt\beta B_\rho]$, $W_\rho=M_\rho M_\rho^\top=G+E_\rho\succ0$.
The shape has exact diagonal $\diag(G)$, paired $|(E_\rho)_{ij}|\le\beta r$, zero elsewhere.
For prescribed joint budget $B\ge0$, the coefficient ball $\|\Theta\|_F\le B$
has image $\{Y:\tr(Y^\top W_\rho^{-1}Y)\le B^2\}$ under $M_\rho$:
the minimum-norm preimage $M_\rho^\top W_\rho^{-1}Y$ has squared norm
$\tr(Y^\top W_\rho^{-1}Y)$.

Verified $G\succeq\mu_G I$ gives $\delta=\beta r/\mu_G$; paired support gives
$\|E_\rho\|_{\op}\le\beta r$ (Frobenius bound $\sqrt n\beta r$).
Since $W_\rho\preceq(1+\delta)G$, a squared-radius upper bound $U_S$ gives
\[
 \frac{\|A_ZY-\widehat Y^{(S)}\|_F}{\sqrt q}
 \le B\sqrt{(1+\delta)U_S/q}.
\]
For $\delta<1$, fixed-subset inflation against canonical decoding is
$\le(1-\delta^2)^{-1/2}$. Score-vector RMS divides by $\sqrt q$
(benchmark fidelity: $\sqrt{qc}$). Among queries with reference margin
$\ge\gamma>0$, RMS $\le t$ allows $\le2qt^2/\gamma^2$ argmax changes,
each requiring squared row error $\ge\gamma^2/2$; no margin distribution is assumed.

A post-review design fixes $B=1$, six cosine caps $0$--$0.5$ and RMS policies
$0.01,0.02,0.05,0.10$ before aggregating response-free decisions for $120$ stored
text geometries and eight methods. Exact-rational source checks cover $24{,}480$
decisions, including $1{,}440$ for $60$ random-subset cases.
\begin{table}[t]
\centering
\begin{tabular}{lrr}
\toprule
Target & $r=0$ & $r=0.25$\\
\midrule
Original & $0/30$ & $0/30$\\
No constant & $0/30$ & $0/30$\\
Query contrast & $30/30$ & $20/30$\\
Centered kernel & $30/30$ & $20/30$\\
\bottomrule
\end{tabular}
\caption{Constrained-head certifications of score-vector root mean square (RMS)
error at most $0.05$. Counts coincide for power four and adaptive direction;
nonapproval does not imply impossibility.}
\label{tab:application-contract}
\end{table}
On twelve first-seed geometries, both selectors approve $6/12$, $4/12$ at
$r=0,0.25$; five random subsets per geometry give $6/60$, $0/60$.
At $r=0.05,0.10,0.25$, entrywise bounds approve $5\%$ inflation for all $120$
geometries; Frobenius approves none. Benchmark-label/large-language-model head
membership and heldout-calibrated accuracy remain unestablished.

Each head trains on all $512$ labels. Scaling/rounding verify membership on
twelve geometries; refits read $128$ teacher-score vectors. Separate lower-upper
(LU) solves share the enclosure library. Residual checks validate $156$ endpoints
(RMS $0.00938$--$0.01822$); zero-predictor RMS $0.02337$--$0.04446$ makes $0.05$ trivial.
Post hoc exact comparisons of outward enclosures put all $156$ errors
(eight methods, five random controls) below the existing $0.02$ policy;
exact stored-prediction squared norms and residual allowances put all twelve
zero-predictor errors above it.
For fixed stored $Y$, put $T=A_ZY$, $T_S=\widehat Y^{(S)}$.
For stored full prediction $P$, let $e\ge\|T-P\|_F$,
$s_-\le\|P\|_F$, $r_+\ge\sqrt q$, and
$u_S\ge\|T-T_S\|_F^2/q$. If $s_->e$, reverse triangle gives
\begin{equation}
 \frac{\|T-T_S\|_F}{\|T\|_F}
 \le\frac{\sqrt{u_S}}{\ell},\qquad
 \ell=\frac{s_--e}{r_+}>0.
 \label{eq:fixed-response-relative-enclosure}
\end{equation}
Exact dyadic squared norms and directed rational square roots give $s_-,r_+$.
Use the larger of the two retained uppers for $u_S$;
exact $u_S<a^2\ell^2$ proves ratio $<a$ for $a>0$.

Under the declared arithmetic model, paired bounds in Equation~\ref{eq:fixed-response-relative-enclosure}
certify relative error $<0.622$ for all $156$ refits, $<0.422$ for power four.
These fixed-response
stored-matrix bounds establish neither class-uniform guarantees at $0.02$ or
in relative error nor selector superiority.

Fix $S$, $A_Z\ne0$. On a full-dimensional centered ellipsoid, restrict relative
error to $A_ZY\ne0$. If $A_ZE_{S^c}\ne0$, $S^c=[n]\setminus S$, opposite
hidden responses force minimax relative error one over $Y_S$-only decoders
(Lemma~\ref{lem:floor}); zero attains it. Otherwise recovery is exact.
This covers the full head class at fixed $\rho$, $B>0$; further head restrictions
require separate analysis.

In each geometry, verified directional energies and Cauchy--Schwarz give a unit
$K_\beta$-head with zero-predictor RMS above $0.05$. Zero calibration coefficients
ensure membership in every allowed $W_\rho$ class. Table~\ref{tab:application-contract}
certifies all class responses at $0.05$ for both selectors: $60/120$ at $r=0$,
$40/120$ at $r=0.25$. Witnesses may vary by geometry.

\subsection{Earlier Observed Numerical Tasks}\label{sec:observed}
An earlier comparison uses scikit-learn 1.8.0 Diabetes, Breast Cancer Wisconsin
(Diagnostic), Wine and Digits \citep{sklearn_toy180}. Twenty input-only
partitions per task separate reference/query pools and group exact duplicates;
$n=256,q=128$, except Wine ($n=96,q=64$). Reference-only preprocessing,
linear and radial basis function (RBF) kernels with a constant term, three ridge
shifts and three budgets are frozen before outcomes. All 480 geometry/ridge
cases and 20,160 evaluations across 14 arms remain archived, including the
three generalized leverage score (GLS) presets of Section~\ref{sec:text-extensions}.

At primary RBF, $\alpha=10^{-3},m=n/4$, powers two/four reduce radius by
$2.7$--$11.2\%$/$2.9$--$16.8\%$, but seven of eight fidelity intervals include
one. Digits accuracy is $90.0\%$ for $p=1,2$, $89.1\%$ for $p=4$, and
$93.4\%$ for full KRR. The post-protocol Holm sensitivity rejects all eight
radius nulls, without making the fixed tasks independent populations.
Generic score bounds at $k=n/4$ are zero; none of the 1,920 evaluated bounds
at $k=n/16,t=3n/4$ improves rank removal. Later direction/rank/overlap
refinements also fail to improve the combined bounds. All 223,680 overlap
branches and adverse outcomes remain as floating diagnostics.
Appendix~\ref{app:numerical-implementation} states historical arithmetic
and timing conventions; this study establishes no training acceleration.

\section{Subset Refitting on Frozen Text Embeddings}
\label{sec:text-embeddings}
Frozen pretrained representations supply KRR inputs without fine-tuning;
benchmark annotations supply responses. Radius, fidelity and task performance
are evaluated separately.

\subsection{Data, Representations, and Information Boundaries}
\label{sec:text-data}
BANKING77 has 77 intent classes \citep{casanueva2020banking}; AG News and
DBpedia14 have four news and fourteen ontology classes \citep{zhang2015text}.
Official splits and original texts are preserved. Duplicate keys use Unicode
Normalization Form KC (NFKC), collapsed whitespace and case folding: cross-split groups are excluded
on both sides; within-split groups retain one input-hash-ordered representative.
This removes normalized duplicates, not semantic near-duplicates
(Table~\ref{tab:text-inputs}).

\begin{table}[t]
\centering\small
\setlength{\tabcolsep}{5pt}
\begin{tabular}{lrrrr}
\toprule
Data set & Classes & Eligible train & Eligible test & Embedded texts\\
\midrule
BANKING77 & 77 & 9,992 & 3,072 & 11,864\\
AG News & 4 & 119,839 & 7,590 & 22,575\\
DBpedia14 & 14 & 559,999 & 70,000 & 23,705\\
\bottomrule
\end{tabular}
\caption{Inputs after duplicate removal. Embedded texts unite reference pools,
1,024 pilot inputs, 512 selection queries and 2,048 heldout queries per corpus.
The 116,288 vectors represent 58,144 distinct texts in two models.}
\label{tab:text-inputs}
\end{table}

The primary \texttt{text-embedding-3-large} and replication
\texttt{text-embedding-3-small} representations have 3,072 and 1,536 coordinates
\citep{openai2024embeddings}. Identical untruncated, row-ID-aligned texts are
requested without labels. Returned float32 vectors enter float64 kernel
computations without further normalization. Across 912 batches, model IDs,
dimensions, byte/input hashes, request metadata and usage agree internally,
without independent attestation or pinned backend versions. Changing model
and dimension together does not isolate model size.

Ten input-only draws per corpus use seeds 720001--720010 and nested
$n=512,1024,2048$, covering 8,280/18,991/20,121 training documents in corpus
order. Disjoint roles comprise 1,024 training pilots, $q=512$ official-test
selection queries and 2,048 evaluation-only test queries per corpus. Query
IDs are reused across draws, budgets and representations. Excluding 3,072
pilot texts leaves 55,072 texts. The transductive radius certifies the fixed
selection queries, not the heldout operator.

Selectors receive vectors and $G,C,R_\varnothing$. Validated subsets for all six
methods and three budgets freeze with input hashes before label release.
Schema validation withholds targets and their distributions in a separate
vault; this trusted contract provides no operating-system isolation.

\subsection{Fixed Analysis and Evaluation}
\label{sec:text-analysis}
A dated local amendment fixes analysis after embedding collection but before
KRR outcomes, without external preregistration. Planned IMDb vectors were
unavailable for verification despite logs reporting success; availability
determined the original three corpora before outcomes opened.

Input gates use BANKING77 large-model pilots at $n=512,1024$ and seed 720001
at $n=2048$. Decisions use only time, memory, admission and backward error;
scores, radii and labels remain closed. All pass 1,200 seconds and 4 GiB;
the largest takes 404.0 seconds and 643.6 MiB. Primary settings are the large
model, $n=2048$, $q=m=512$, constant-plus-RBF kernel and $\alpha=10^{-3}$.

Reference inputs determine coordinate means, population standard deviations
(zeros replaced by one) and median positive squared distance $h^2$.
Use $k(x,z)=1+\exp(-\|x-z\|^2/(2h^2))$, unchanged $\lambda=n\alpha$ and
supplied $W_0=G$. Powers $p=1,2,4$, greedy log-det, ridge-leverage ranking
and nested random ordering use $m/n\in\{1/16,1/8,1/4\}$ and
$C_SG_{SS}^{-1}Y_S$. Targets are uncentered one-hot vectors in the public
vocabulary. Argmax yields classes; macro-F1 uses the full vocabulary, setting
undefined classwise scores to zero. Other models, sizes, budgets, comparators,
task and holdout measures are descriptive.

Primary outcomes are normalized radius
$\bar\rho_S=\sqrt{\lambda_{\max}(R_S)/\lambda_{\max}(R_\varnothing)}$,
where $R_\varnothing=CG^{-1}C^\top$, and query fidelity RMS
$\|\widehat Y_S-\widehat Y_{\rm full}\|_F/\sqrt{qc}$ for $c$ classes.
All reported denominators are positive; zero initial radius has no normalized ratio.
For each corpus and $p=2,4$, all ten unrounded paired log ratios to $p=1$
yield a geometric mean, marginal 95\% Student $t_9$ confidence interval (CI)
and two-sided zero-mean test. Ten draws give limited support for the $t$
approximation. Inference conditions on corpus, representation and queries;
overlapping draws, reused queries and nested designs add no independent corpora.

Holm adjustment at family-wise level $0.05$ retains sixteen planned slots:
four corpora, two powers and two outcomes. Four unavailable IMDb slots have
$p=1$ and no effects; twelve comparisons ran. Undefined log ratios,
constant sequences and nonfinite statistics remain nonestimable with $p=1$,
without epsilon additions or deletions. Intervals are nonsimultaneous.
Separately, 2,000 paired heldout-ID resamples carry each document's $p=1,2,4$
predictions across draws and models together; percentile intervals describe
task-score differences conditional on fitted predictors, without combining
them with draw intervals into population intervals.

\subsection{Radius, Realized Fidelity, and Heldout Performance}
\label{sec:text-results}
All 180 original geometries and 3,240 method/budget outcomes complete without
an omitted case. Powers two/four reduce geometric-mean radius by
2.64--3.03\%/4.39--5.19\% relative to trace; all six comparisons survive
the sixteen-slot Holm adjustment ($p<5\times10^{-9}$).
No fidelity improvement survives that adjustment. Power four instead increases
BANKING77/DBpedia14 query error by 1.46\%/1.51\%, with adjusted
$p=5.1\times10^{-4}/2.7\times10^{-5}$; AG News intervals include equality
(Table~\ref{tab:text-primary}).

Among the original six selectors, $p=4$ has the smallest mean radius on each
corpus. The best query fidelity occurs at $p=2,4,1$ for BANKING77, AG News,
and DBpedia14, respectively (Table~\ref{tab:text-all}).
Task rankings differ (Table~\ref{tab:text-holdout}): BANKING77 $p=4$ increases
mean heldout accuracy from 69.86\% to 72.82\% and macro-F1 from 66.37\% to
70.65\%, with positive conditional document-bootstrap intervals
(Table~\ref{tab:text-holdout-bootstrap}). AG News intervals include zero;
DBpedia14 accuracy decreases for $p=2$ and increases for $p=4$.
Log-det has the highest subset accuracy on AG News and DBpedia14 despite
larger query radius and fidelity error. Full KRR exceeds every subset method's
mean accuracy on all three corpora, using four times as many responses.
Even on the shared BANKING77 selection-query batch, $p=4$ raises macro-F1
while worsening reference fidelity. These descriptive comparisons supply
neither further primary tests nor a general task guarantee.

Across three sizes and two representations at $m=n/4$, all 36 radius ratios
are below one, but 30 fidelity ratios exceed one. The small representation
worsens AG News fidelity for both higher powers at $n=2048$, whereas the
large representation has lower point estimates. Reused inputs do not make
these settings independent replications.

\par\noindent\emph{Post hoc paired-sign sensitivity.}\label{sec:text-sign}
For positive/negative log-ratio counts $b_+,b_-$ and $N=b_++b_-$, the paired-sign
value is $\min\{1,2\Pr[B\le\min(b_+,b_-)]\}$ for
$B\sim\operatorname{Binomial}(N,1/2)$. This post hoc test of independent sign
balance reuses the ten draws; primary mean-log-ratio tests remain unchanged. All 120 pairs
are nonties. All six radius contrasts and BANKING77/DBpedia14 $p=4$ fidelity
increases have raw $p=1/512$ and Holm-adjusted $p=1/32$ across sixteen slots,
including four unavailable IMDb slots at one. Other fidelity contrasts have
adjusted $p=1$; this is not independent replication.

\begin{table}[t]
\centering\small
\setlength{\tabcolsep}{3pt}
\begin{tabular}{@{}llrrrr@{}}
\toprule
Data set & Power & \shortstack{Radius ratio\\(95\% CI)} & Holm $p$ & \shortstack{Fidelity ratio\\(95\% CI)} & Holm $p$ \\
\midrule
BANKING77 & $p=2$ & $0.9736\,[0.9715,0.9757]$ & $4.7\!\times\!10^{-9}$ & $0.9988\,[0.9930,1.0047]$ & $1.000$ \\
BANKING77 & $p=4$ & $0.9561\,[0.9537,0.9584]$ & $1.8\!\times\!10^{-10}$ & $1.0146\,[1.0099,1.0193]$ & $5.1\!\times\!10^{-4}$ \\
\addlinespace[3pt]
AG News & $p=2$ & $0.9702\,[0.9691,0.9712]$ & $4.3\!\times\!10^{-12}$ & $0.9935\,[0.9841,1.0029]$ & $1.000$ \\
AG News & $p=4$ & $0.9482\,[0.9468,0.9497]$ & $7.9\!\times\!10^{-13}$ & $0.9885\,[0.9753,1.0020]$ & $0.681$ \\
\addlinespace[3pt]
DBpedia14 & $p=2$ & $0.9697\,[0.9679,0.9716]$ & $5.1\!\times\!10^{-10}$ & $1.0008\,[0.9964,1.0051]$ & $1.000$ \\
DBpedia14 & $p=4$ & $0.9481\,[0.9461,0.9501]$ & $1.0\!\times\!10^{-11}$ & $1.0151\,[1.0118,1.0184]$ & $2.7\!\times\!10^{-5}$ \\
\bottomrule
\end{tabular}

\caption{Primary text comparisons against $p=1$ at $n=2048,m=512$ with the
large representation. Geometric ratios below one favor the stated power;
brackets give marginal 95\% $t_9$ intervals. Adjusted $p$-values use all
sixteen planned Holm slots, including four unavailable IMDb slots at one.
The complete paired values and raw $p$-values are retained with the results.}
\label{tab:text-primary}
\end{table}

\begin{table}[t]
\centering\small
\setlength{\tabcolsep}{4pt}
\begin{tabular}{@{}lrrr@{}}
\toprule
Method & BANKING77 & AG News & DBpedia14 \\
\midrule
Trace $p=1$ & $0.0445\,/\,0.0241$ & $0.0447\,/\,0.1009$ & $0.0449\,/\,0.0701$ \\
Power $p=2$ & $0.0434\,/\,0.0241$ & $0.0434\,/\,0.1003$ & $0.0435\,/\,0.0702$ \\
Power $p=4$ & $0.0426\,/\,0.0245$ & $0.0424\,/\,0.0998$ & $0.0425\,/\,0.0712$ \\
Log-det & $0.0695\,/\,0.0308$ & $0.0601\,/\,0.1178$ & $0.0532\,/\,0.0831$ \\
Ridge leverage & $0.0817\,/\,0.0362$ & $0.0658\,/\,0.1393$ & $0.0639\,/\,0.1060$ \\
Random & $0.0503\,/\,0.0283$ & $0.0481\,/\,0.1096$ & $0.0483\,/\,0.0766$ \\
\bottomrule
\end{tabular}

\caption{Primary text experiment: mean normalized radius / query fidelity RMS
over ten matched reference draws. All six methods share the same $m=512$
response budget. Entries are conditional on the stated representation and
kernel, and fidelity scales differ across class vocabularies.}
\label{tab:text-all}
\end{table}

\begin{table}[t]
\centering\small
\setlength{\tabcolsep}{4pt}
\begin{tabular}{@{}lrrr@{}}
\toprule
Method & BANKING77 & AG News & DBpedia14 \\
\midrule
Full reference & $87.12\,/\,86.55$ & $90.62\,/\,90.70$ & $97.61\,/\,97.59$ \\
\addlinespace[3pt]
Trace $p=1$ & $69.86\,/\,66.37$ & $88.35\,/\,88.37$ & $92.81\,/\,92.42$ \\
Power $p=2$ & $72.14\,/\,69.29$ & $88.41\,/\,88.43$ & $92.48\,/\,92.08$ \\
Power $p=4$ & $72.82\,/\,70.65$ & $88.52\,/\,88.55$ & $93.75\,/\,93.48$ \\
Log-det & $64.17\,/\,58.64$ & $89.31\,/\,89.35$ & $95.87\,/\,95.76$ \\
Ridge leverage & $52.39\,/\,44.69$ & $87.54\,/\,87.44$ & $77.60\,/\,74.48$ \\
Random & $70.04\,/\,66.61$ & $88.95\,/\,89.00$ & $94.94\,/\,94.74$ \\
\bottomrule
\end{tabular}

\caption{Heldout accuracy / macro-F1 in percent, averaged over ten reference
draws at the primary setting. Full KRR uses all 2,048 responses; the six
subset methods use 512. The 2,048 heldout documents per data set were excluded
from selection. These task results are descriptive, not consequences of the
query-radius theorem.}
\label{tab:text-holdout}
\end{table}

\par\noindent\emph{An evaluator-only error decomposition.}
For a descriptive decomposition, set $D_S=CG^{-1}-C_SG_{SS}^{-1}E_S^\top$,
$\rho_S=\sqrt{\lambda_{\max}(R_S)}$, and
$E_Y=\tr(Y^\top G^{-1}Y)$. Compatibility gives $R_S=D_SGD_S^\top$.
For $\rho_S\sqrt{E_Y}>0$, define the evaluator-only utilization
\begin{equation}
 \upsilon_S=\frac{\|D_SY\|_F}{\rho_S\sqrt{E_Y}}\in[0,1],
 \qquad
 f_S=\frac{\|D_SY\|_F}{\sqrt{qc}}
     =\rho_S\sqrt{\frac{E_Y}{qc}}\,\upsilon_S.
 \label{eq:text-utilization}
\end{equation}
For positive quantities in the same geometry,
$f_S/f_{S_1}=(\rho_S/\rho_{S_1})(\upsilon_S/\upsilon_{S_1})$.
This prespecified algebraic diagnostic uses all responses, never selection;
it supplies no causal explanation. Zero denominators remain undefined.
Floating values remain unclipped with their declared allowance; $[0,1]$ is an
exact-arithmetic bound. Section~\ref{sec:application-contract} supplies a
decision from an independently prescribed response budget.
At the primary setting, the 90 utilization values for $p=1,2,4$ across the
three corpora and ten reference draws range from 0.1537 to 0.2693.
The geometric utilization ratios relative to $p=1$ are 1.0240--1.0320 for
$p=2$ and 1.0425--1.0707 for $p=4$. Thus the radius reduction accompanies
greater utilization of the remaining envelope by the observed responses.
For BANKING77 at $p=4$, the rounded factors give
$0.9561\times1.0612\approx1.0146$; for DBpedia14 they give
$0.9481\times1.0707\approx1.0151$. In both cases, the utilization increase more
than offsets the smaller radius. The decomposition quantifies the observed
fidelity gap while retaining the supplied shape $G$ and the unchanged decoder.

\par\noindent\emph{Scope and computational cost.}
The complete selection stage takes 99.3 minutes with eight concurrent workers;
evaluation takes a further 5.3 minutes with three workers. No case fails or
times out. At $n=2048$ with the large representation, median selector-path
wall times across the 30 corpus/seed cases are 132.1, 164.7 and 188.2 seconds
for $p=1,2,4$, respectively, compared with 125.7 seconds for log-det,
1.04 seconds for ridge-leverage ranking and less than 0.001 seconds for
random ordering. These times include checks performed inside the selector paths
and reflect concurrent execution. Shared geometry setup and independent
post-path Schur checks are included in the complete selection time but excluded
from these method-path times; subsequent evaluation is recorded separately.
The largest recorded selection-worker peak is 763.9 MiB. Dense matrix
construction, admission and selection remain material costs; a smaller
response budget alone does not establish a computational speedup.
The original primary study uses one kernel construction and one value of
$\alpha$. Section~\ref{sec:primary-certificates} adds post hoc lower-bound
diagnostics on its unchanged geometries and subsets.

These cached-vector results imply neither generalization across embedding
models nor contamination-free evaluation, training or deployment savings.
All reference labels formed the full-data targets; retained budgets do not
measure realized annotation savings.

\subsection{Verified Same-Budget Bounds}\label{sec:primary-certificates}
\begin{table}[!htb]
\centering\small
\setlength{\tabcolsep}{4pt}
\begin{tabular}{lrrrrrr}
\toprule
Corpus & $p=1$ & $p=2$ & $p=4$ & Fixed direction & Exact $p=2$ & Exact $p=4$ \\
\midrule
BANKING77 & 1.114228 & 1.080287 & 1.060764 & 1.053235 & 1.082284 & 1.061069 \\
AG News & 1.108604 & 1.075487 & 1.051744 & 1.045003 & 1.074468 & 1.049669 \\
DBpedia14 & 1.108233 & 1.072990 & 1.045974 & 1.040766 & 1.071165 & 1.045051 \\
\bottomrule
\end{tabular}

\caption{Largest verified returned-to-optimal radius-ratio upper bound over
all ten original primary draws per corpus. Displayed bounds round upward.
The denominator bounds the optimum at the same $m=512$ response budget;
Appendix~\ref{app:verified-arithmetic} states the exact matrix target and
arithmetic assumptions.}
\label{tab:primary-certificates}
\end{table}

At $q=k=m=512$, rank removal is zero. Generic witnesses at prefixes
$0,128,256,512$ give positive floating radius ratios $9.5$--$25.1$; identity
attains every positive maximum, and overlap cannot improve the rank ceiling.
Even with $\gamma_0=1$ and $\epsilon=0.1v_p(\varnothing)$, sufficient $p=2,4$
budgets are $2{,}358$ and $4{,}716$, both above $n=2048$; these are not necessary label counts.

Proposition~\ref{prop:krr-directional} uses the leading $R_\varnothing$
direction without labels or tuning. Appendix~\ref{app:verified-arithmetic}
encloses kernel energy and radii for stored $G,C,\lambda$, verifying
$G-\lambda I\succ0$ under the stated binary64 and Basic Linear Algebra Subprograms (BLAS) model. All 30 geometries and 180 rows pass positivity gates. Corpus maxima in
Table~\ref{tab:primary-certificates} are below 1.081 ($p=2$) and 1.061 ($p=4$).
These selector-independent radius certificates exclude ideal unrounded kernels
and realized fidelity. Post hoc $p=4$ mean relative score errors
$\|\widehat Y_S-\widehat Y_{\rm full}\|_F/\|\widehat Y_{\rm full}\|_F$
are $43.0\%,25.0\%,39.0\%$ in corpus order despite near-optimal radii.

With prior-source/network access denied, all 30 snapshots and six subsets
per geometry replay; 180 bounds match bitwise. The largest squared-radius
operator-error allowance is $3.079\times10^{-6}$. Mean replay cost is
6.37 seconds per geometry under shared load, including loading/verification
and excluding geometry/selection; no speed comparison follows. The directional
bound followed 17 generic cases; post hoc origin, unenclosed estimates,
versions and failures remain recorded.

\FloatBarrier

\section{Original-Size Directional Controls}\label{sec:directional-controls}
The original $1+\mathrm{RBF}$ geometries have
$\lambda_{\max}(R_\varnothing)/\tr(R_\varnothing)=0.8954$--$0.9351$;
their initial leading direction retains $0.9681$--$0.9947$ of the returned
top residual eigenvalue on the $p=2,4$ paths. Fixed-direction greedy (FD)
fixes a leading unit vector $v$ of $R_\varnothing$ and maximizes
$(v^\top b_i)^2/Q_{ii}$ at every step. The exploratory original-size comparison
adds FD and exact-$2/4$ to all 30 retained $n=2048,q=m=512$ geometries.
Exact marginal denotes an objective formula evaluated in floating arithmetic.
Dated protocols disclose the motivating diagnostic and preserve every case,
interrupted attempt and source version; the original sixteen slots are unchanged.

\begin{table}[!htb]
\centering\small
\setlength{\tabcolsep}{3pt}
\begin{tabular}{llrrrr}
\toprule
Corpus & Ratio & Radius [95\% CI] & Query RMS [95\% CI] & Holdout & $\Delta$F1 \\
\midrule
BANKING77 & FD/$p=4$ & 0.9906 [0.9889, 0.9923] & 1.0233 [1.0145, 1.0322] & 1.0108 & +0.95 \\
BANKING77 & Exact $2$/$p=2$ & 1.0017 [1.0001, 1.0033] & 1.0030 [0.9989, 1.0070] & 1.0008 & -0.06 \\
BANKING77 & Exact $4$/$p=4$ & 0.9995 [0.9988, 1.0003] & 1.0083 [1.0022, 1.0144] & 1.0062 & -0.14 \\
AG News & FD/$p=4$ & 0.9944 [0.9935, 0.9954] & 1.0167 [1.0025, 1.0311] & 1.0015 & +0.47 \\
AG News & Exact $2$/$p=2$ & 0.9991 [0.9977, 1.0005] & 1.0006 [0.9926, 1.0087] & 0.9997 & +0.01 \\
AG News & Exact $4$/$p=4$ & 0.9990 [0.9984, 0.9996] & 1.0037 [0.9911, 1.0166] & 0.9996 & +0.12 \\
DBpedia14 & FD/$p=4$ & 0.9936 [0.9931, 0.9942] & 1.0252 [1.0212, 1.0291] & 1.0169 & +1.93 \\
DBpedia14 & Exact $2$/$p=2$ & 0.9985 [0.9971, 1.0000] & 0.9979 [0.9941, 1.0017] & 0.9993 & +0.23 \\
DBpedia14 & Exact $4$/$p=4$ & 0.9988 [0.9984, 0.9993] & 0.9988 [0.9963, 1.0012] & 0.9976 & +0.42 \\
\bottomrule
\end{tabular}

\caption{Original-size geometric-mean ratios ($n=2048,m=512$); below one
favors the numerator. Holdout is heldout fidelity RMS; $\Delta$F1 is the mean
macro-F1 difference in percentage points. Intervals are marginal; complete
paired outcomes and heldout intervals remain in the artifact.}
\label{tab:primary-directional}
\end{table}

FD has smaller radius in all 30 cases: corpus means are $0.56$--$0.94\%$
below $p=4$, with $1.67$--$2.52\%$ worse query fidelity but $0.47$--$1.93$
points higher heldout macro-F1 (Table~\ref{tab:primary-directional}). Exact-$4$
radius gains are $0.05$--$0.12\%$; BANKING77's interval crosses one and its
query fidelity worsens by $0.83\%$. Tight certification therefore establishes
neither power-selection superiority nor realized prediction closeness.

A separate matched-cost panel reruns all six methods on the first draw per
corpus, under fixed rotated order and single-thread arithmetic. All 18 calls
recover their subsets. Selection central processing unit (CPU) time is
$87.12$--$97.73$ seconds for $p=4$ and $22.35$--$23.52$ for FD; matched
FD/$p=4$ call ratios, including initialization and validation, are
$0.238$--$0.263$. Exact/proxy ratios are $1.016$--$1.189$ for $p=2$ and
$1.042$--$1.255$ for $p=4$. These shared-host, single-call costs exclude
geometry and prediction and establish no end-to-end speedup.

The earlier centered/contrast/kernel controls and later same-size addendum
retain all 480 new paths and 660 evaluations, with no numerical failures.
Independent code checks all saved prediction metrics and 66 endpoints;
scaled prediction and residual-spectrum discrepancies are at most
$6.79\times10^{-14}$ and $1.45\times10^{-14}$. Their complete records remain;
Section~\ref{sec:multidirectional-study} extends the same-size comparisons
to all eight methods on all four targets.

\subsection{The Original-Scale Adaptive Extension}
\label{sec:scale-extension}
A frozen exploratory extension runs adaptive direction (AD), as defined in
Section~\ref{sec:multidirectional-study}, unchanged on all thirty original
$n=2048$, $q=m=512$ geometries. Thirty new paths plus six retained
control methods yield 210 rows, with unchanged inputs, ridge, bandwidth and queries.
This increases reference-pool scale, not independent corpora or representations.

All thirty paths and ninety new adaptive/power-four/fixed-direction certificate
comparisons complete. Adaptive direction improves radius in 30/30 cases against
power four and 26/30 against fixed direction (Table~\ref{tab:scale-extension}).
Query fidelity worsens in every corpus mean; heldout fidelity improves only
for BANKING77/AG News. Macro-F1 improves against power four in all three means.
These are descriptive results without new primary tests.

Nine first-seed timing calls reproduce scientific sets. Adaptive/power-four
CPU ratios are $0.467$--$0.511$, versus $1.752$--$1.927$ for adaptive/fixed
direction. Selector initialization/validation are included, geometry,
prediction and certification excluded. Single shared-host calls establish
no population or end-to-end savings.

\begin{table}[t]
\centering\small
\begin{tabular}{lrrrrrr}
\toprule
Data set & Radius & Query & Heldout & $\Delta$F1 & AD/FD & Bound\\
\midrule
BANKING77 & 0.9845 & 1.0086 & 0.9976 & $+1.209$ & 0.9938 & 1.045\\
AG News   & 0.9937 & 1.0062 & 0.9902 & $+0.468$ & 0.9993 & 1.044\\
DBpedia14 & 0.9925 & 1.0168 & 1.0069 & $+1.752$ & 0.9989 & 1.039\\
\bottomrule
\end{tabular}
\caption{Original-scale extension. Radius, query and heldout columns give
ten-draw geometric-mean AD/power-four ratios; AD/FD compares radius with fixed
direction. $\Delta$F1 is mean heldout macro-F1 difference to power four in
percentage points. Bound is the largest verified adaptive-to-optimal
same-budget radius-ratio upper bound, rounded upward.}
\label{tab:scale-extension}
\end{table}

The original directional witness certifies every adaptive subset within 1.045
of its same-budget optimum for exact stored matrices. A frozen first-seed
search on all three corpora uses ranks $1,4,16$ and sixteen conditional-gradient
steps per rank. All three verified variational lowers lose to the original
directional lower; their maximum therefore retains it. Extra directions need
not improve an informative common-direction certificate.

Independent canonical-input reconstruction checks thirty new endpoints using
LU refits/direct residuals (maximum scaled discrepancy $3.70\times10^{-14}$),
and reads back 180 retained score rows. These floating comparisons are separate
from outward certification.

\section{Earlier Exploratory Extensions}\label{sec:text-extensions}
Earlier cached-input extensions retain 429 six-method geometries (2,574
outcomes), 120 stronger-comparator outcomes, and all protocols, source versions,
failures and timings. They add no original hypothesis slots or independent
corpora. Their complete tables remain in the artifact.

\par\noindent\emph{Sensitivity and alternative selectors.}
At $n=512,m=128$, the ridge/bandwidth grid gives 54 power/trace mean-radius
ratios below one, but query fidelity improves in only 12/27 settings for
$p=2$ and 6/27 for $p=4$. Representation truncation, new cached-pool draws
and nested-size comparisons likewise give radius gains without uniform
fidelity gains. A 900-second $n=4096$ input gate fails; its $n=2048$ fallback
does not extend the original maximum size. Each changed kernel defines a
different reference and supplied shape.

The generalized leverage score (GLS) presets of \citet{ordozgoiti2022} retain
dictionary directions covering $75\%$ of squared Frobenius norm, rank them by
target energy, and use fractions $0.10,0.25,0.50$ of retained rank (rounded up).
For $G^{1/2}$ and $G^{-1/2}C^\top$, direction $(\kappa_j,u_j)$ has energy
$\|Cu_j\|_2^2/(\kappa_j+\lambda)$; the largest $m$ row leverages select labels.
This fixed-budget adaptation is basis-dependent in repeated eigenspaces and
does not invoke the published stopping theorem. GLS has worse mean radius and
query fidelity on all three original text targets. An ordinary-arithmetic
expected-polynomial prototype fails root admission at $n=16$; this neither
implements the certified algorithm of \citet{caixuxu2026} nor establishes its
infeasibility.

\par\noindent\emph{Exact marginals and dependence.}\label{sec:exact-ablation}
The earlier 180-path exact-marginal comparison is exploratory, including its
dated exact-$4$ addendum. Mean radius gains are small and all six marginal
intervals include one; exact-$4$ worsens BANKING77 query fidelity in every
draw. Exact/proxy final $\tr(R_S^4)$ ratios are $1.0520,1.0146,0.9979$ in
BANKING77/AG News/DBpedia14 order: exact one-step gains do not order final
objectives. The new matched study below extends these comparisons.
Similarity masks and paired fixed-fit resampling retain all documents or
declared exclusions; they establish neither source independence nor absence
of pretraining overlap. Their complete conditional intervals are archived.

\par\noindent\emph{Alternative responses and late-available IMDb.}
\label{sec:teacher-responses}\label{sec:imdb-completion}
The 21,745 cached \texttt{gpt-5.4-2026-03-05} classifications collected on
November 19, 2026 yield 540 outcomes. Responses load only after subsets freeze;
unchanged radius is therefore a design consequence. Only 1/18 query and 4/18
heldout power/trace fidelity ratios improve. One generation per document and
client-retained records do not establish generation stability or provider
attestation. The recovered IMDb study \citep{maas2011imdb} adds 60 selection
cases and 1,080 outcomes without new requests: all 36 radius ratios improve,
but only five query and six heldout fidelity ratios do. At $n=2048,m=512$,
large-representation macro-F1 is $95.95\%$ for full KRR, $95.16\%$ for trace,
$94.83\%$ for $p=4$, and $95.25\%$ for random. These descriptive late results
leave the sixteen original slots unchanged. The four-corpus union contains
75,827 distinct documents; representations and teacher responses add none.
\FloatBarrier

\section{Matched Multidirectional Comparisons}
\label{sec:multidirectional-study}
The post-review grid freezes 120 cases: three corpora, ten draws, four targets
at $n=q=512,m=128$ (original, no constant, query contrast, reference-centered).
Retained trace, power-$2/4$ and fixed-direction paths are joined by exact-$2/4$,
spectral greedy (minimizing the next largest residual eigenvalue) and adaptive
direction (maximizing decrease along the updated leading direction).
All eight share geometry, ties and budget; adaptation uses only residual
geometry. Labels and heldout vectors remain closed until selection freezes.
This exploratory design adds no primary tests.

With $H_d=I_d-\one_d\one_d^\top/d$, contrast replaces $C$ by $H_qC$;
no constant uses RBF alone. Centering the RBF reference gives
$K_c=H_nKH_n$ and
\[
 C_c=C-\frac{(C\one_n)\one_n^\top}{n}
       -\frac{\one_q(\one_n^\top K)}{n}
       +\frac{\one_n^\top K\one_n}{n^2}\one_q\one_n^\top.
\]
Heldout kernels use reference means. Ridge, bandwidth and order are fixed,
without a fitted intercept. Controls change the reference; kernel changes
also change the shape. Comparisons remain within target; contrast has no
classification interpretation.

\begin{table}[!htb]
\centering\small
\setlength{\tabcolsep}{3.5pt}
\begin{tabular}{llrrrrrrr}
\toprule
Target & Corpus & $4/1$ & $p=2$ & FD & Exact $2$ & Exact $4$ & SG & AD \\
\midrule
Original & BANKING77 & 0.9442 & 1.0230 & 0.9881 & 1.0222 & 0.9992 & 1.0011 & 0.9810 \\
Original & AG News & 0.9341 & 1.0265 & 0.9954 & 1.0231 & 0.9990 & 0.9975 & 0.9955 \\
Original & DBpedia14 & 0.9328 & 1.0245 & 0.9954 & 1.0214 & 0.9989 & 0.9958 & 0.9945 \\
No constant & BANKING77 & 0.9496 & 1.0224 & 0.9890 & 1.0233 & 0.9983 & 0.9995 & 0.9807 \\
No constant & AG News & 0.9353 & 1.0246 & 0.9936 & 1.0218 & 0.9966 & 0.9940 & 0.9939 \\
No constant & DBpedia14 & 0.9340 & 1.0250 & 0.9922 & 1.0216 & 0.9975 & 0.9936 & 0.9931 \\
Centered & BANKING77 & 0.9296 & 1.0237 & 1.9035 & 1.0235 & 0.9860 & 1.1709 & 0.9632 \\
Centered & AG News & 0.8951 & 1.0488 & 1.4912 & 1.0389 & 1.0000 & 1.1301 & 0.9663 \\
Centered & DBpedia14 & 0.8849 & 1.0522 & 1.4719 & 1.0391 & 0.9940 & 1.1096 & 0.9651 \\
Contrast & BANKING77 & 0.9272 & 1.0332 & 1.7365 & 1.0300 & 0.9877 & 1.1865 & 0.9698 \\
Contrast & AG News & 0.9036 & 1.0439 & 1.3655 & 1.0357 & 0.9907 & 1.1490 & 0.9621 \\
Contrast & DBpedia14 & 0.8956 & 1.0473 & 1.5591 & 1.0418 & 0.9935 & 1.1185 & 0.9613 \\
\bottomrule
\end{tabular}

\caption{Matched radius comparisons ($n=q=512,m=128$): ten-draw geometric
means. Column $4/1$ compares proxy four with trace; every remaining column
compares its method with proxy four. SG is direct spectral greedy; AD is
adaptive direction. Ratios below one favor the numerator. Full paired values
and marginal intervals are retained; rounded 1.0000 need not be an exact tie.}
\label{tab:multidirectional-radius}
\end{table}

All 480 new paths and 960 method rows complete without numerical failure.
Adaptive direction lowers radius against $p=4$ in 119/120 cases and all twelve
corpus/target means (ratios $0.9613$--$0.9956$;
Table~\ref{tab:multidirectional-radius}). It wins all 60 centered/contrast cases, where
fixed direction loses throughout. Direct spectral greedy instead has larger
radius in all 60, with mean ratios $1.1096$--$1.1866$; exact-$4$ gains range
from negligible to $1.40\%$. Thus proxy four is not favored as a radius default.

Adaptive direction's query/heldout fidelity ratios exceed one in every
corpus/target mean: $1.0075$--$1.0405$/$1.0046$--$1.0247$.
Its centered-target macro-F1 differences are $-3.23,-0.37,-0.51$ points
in BANKING77/AG News/DBpedia14 order. Exact marginals have mixed fidelity;
direct spectral greedy worsens mean query and heldout fidelity in every panel.
Better radius can accompany worse prediction/classification;
intervals are descriptive and nonsimultaneous.

All 96 frozen timing calls reproduce scientific sets. Adaptive/proxy-four
selector CPU ratios are $0.434$--$0.708$ across 12 panels, including
initialization/validation, excluding geometry/prediction/certification.
Single shared-host calls establish no population or end-to-end savings.

\begin{table}[!htb]
\centering\small
\setlength{\tabcolsep}{4pt}
\begin{tabular}{lrrrrrrrr}
\toprule
Target & $p=1$ & $p=2$ & $p=4$ & FD & Exact 2 & Exact 4 & SG & AD\\
\midrule
Original & 1.264 & 1.216 & 1.179 & 1.176 & 1.214 & 1.179 & 1.179 & 1.174 \\
No constant & 1.277 & 1.218 & 1.190 & 1.179 & 1.217 & 1.186 & 1.185 & 1.180 \\
Query contrast & 1.458 & 1.343 & 1.283 & 2.362 & 1.337 & 1.273 & 1.571 & 1.241 \\
Centered kernel & 1.455 & 1.338 & 1.311 & 2.619 & 1.341 & 1.296 & 1.581 & 1.258 \\
\bottomrule
\end{tabular}

\caption{Largest verified same-budget radius-ratio upper bounds across all
30 geometries per target, rounded upward to three decimals. Every one of the
960 method rows verifies for the exact stored matrices. Methods and budgets
match Table~\ref{tab:multidirectional-radius}; the denominator is a lower bound on the
optimum, not its computed value.}
\label{tab:multidirectional-certificates}
\end{table}

\par\noindent\emph{Certification beyond the common direction.}
Witness ranks are $2^j$, $j=0,\ldots,7$: 64 conditional-gradient steps
through rank 64 and the initial rank-128 candidate. Two disclosed BANKING77
development cases preceded the grid freeze. Split
$\beta=\operatorname{fl}(0.99\lambda)$ changes only the certificate. All 120 exact-matrix positivity and positive-bound
gates pass (Table~\ref{tab:multidirectional-certificates}). Power four is certified within
1.311 of its same-budget optimum; adaptive direction within 1.258.
Across all 60 centered/contrast cases, the variational lower bound is
$2.65$--$8.41$ times its within-run verified kernel-energy directional companion;
it wins 95/120 overall. This companion differs from the earlier query-Gram
diagnostic, and no uniform dominance is claimed. Mean per-geometry CPU cost
is 5.27 seconds for geometry, witness proposals and lower verification, plus
1.81 seconds for all eight upper bounds, excluding selection.

Independent code checks all 960 prediction rows and reconstructs twelve
geometries/96 endpoints using lower-upper (LU) factorization and direct
residual operators. Maximum scaled prediction/spectrum discrepancies are
$1.40\times10^{-14}/1.45\times10^{-14}$; all 90 comparable historical
exact/spectral orders match. A separate stored-matrix replay verifies all
120 geometries/960 bounds. Adverse outcomes, admission rules and corrections remain recorded.
\FloatBarrier

\subsection{Calibration Against Full Convex Design}
\label{sec:certificate-comparison}
For the exact fractional proposal in Section~\ref{sec:split-certificate},
write $\mathcal P_k=\{z\in[0,1]^n:\sum_i z_i=k\}$ and
$g_{\rm FW}(z)=\max_{s\in\mathcal P_k}\langle z-s,\nabla f(z)\rangle$.
Since $X(z)=\operatorname{diag}(z)U/\beta$,
\begin{equation}
 L_k^{\rm var}(F,X(z))
 =\frac{[f(z)-g_{\rm FW}(z)]_+}{\|F\|_F^2}.
 \label{eq:fw-identity}
\end{equation}
This is the Frank--Wolfe affine minorant \citep{jaggi2013}; arbitrary $X$
removes reliance on optimizer convergence or exact fractional feasibility.
The preimage proof extends to fractional
$z$ by replacing $U_S$ with $D_zU$, so the same witness lower-bounds the
continuous spectral optimum.

When $K_\beta\succ0$, set $H=CK_\beta^{-1}$,
$B_0=CG^{-1}C^\top-CK_\beta^{-1}C^\top$, and
$M(z)=K_\beta^{-1}+\operatorname{diag}(z)/\beta$.
Woodbury and Schur reduction give the semidefinite program (SDP), adapting
\citet[Proposition~3]{gama2020} from a trace to a spectral objective:
\begin{equation}
 \min_{z\in\mathcal P_k,t} t
 \quad\text{subject to}\quad
 \begin{pmatrix}tI-B_0&H\\H^\top&M(z)\end{pmatrix}\succeq0.
 \label{eq:spectral-relaxation}
\end{equation}
The inverse representation requires this stricter split assumption;
Proposition~\ref{prop:split-certificate} itself permits singular $K_\beta$.

\begin{table}[!htb]
\centering\small
\begin{tabular}{lrrr}
\toprule
$(n,k)$ & $L_{\rm SDP}/L_{\rm fixed}$ &
Max.\ $U_{\rm frac}/L_{\rm SDP}-1$ & CPU ratio\\
\midrule
$(12,3)$ & $1.010$--$1.299$ & $8.18\!\times\!10^{-5}$ & $1.63$--$2.56$\\
$(48,12)$ & $1.016$--$1.246$ & $9.21\!\times\!10^{-6}$ & $339$--$705$\\
\bottomrule
\end{tabular}
\caption{Full spectral relaxation versus fixed-direction variational search:
ranges across twelve cases per size, rounded outward. CPU includes proposal
generation and verification, excluding shared geometry and exhaustive
enumeration. Single shared-host calls do not establish population speed ratios.}
\label{tab:certificate-comparison}
\end{table}

A frozen exploratory comparison takes the first $n=q=12,48$ coordinates/queries
from all three corpora and four targets at seed 720001 in the stored $n=512$
grid: 24 reused-input subgeometries with $k=n/4$, unchanged parent ridge/split,
and no responses read.
The fixed-direction comparator follows Section~\ref{sec:multidirectional-study}'s
rank/iteration schedule, truncated to $q$. CVXPY/Clarabel proposes the full-relaxation weights and query
density. Both lower bounds use the unchanged arbitrary-$X$ verifier.
Strictly feasible dyadic weights and an enclosed residual solve upper-bound
the continuous optimum by $U_{\rm frac}$; solver objectives are diagnostics.
All 24 comparisons verify.

Full optimization strengthens every lower bound, with a substantial cost
increase at $n=48$ (Table~\ref{tab:certificate-comparison}).
For all twelve $n=12$ cases, enclosing every one of the 220 subsets brackets
the discrete squared-radius optimum to width at most $4.50\times10^{-12}$.
The discrete-to-fractional squared-radius optimum ratio lies within $[1.2338,1.5954]$
across these cases: even optimized directions retain a discrete relaxation gap.
The restricted witness trades tightness for cheap, independently checkable
candidates; neither a new design principle nor universal computational
superiority is claimed.

\section{Implications and Limitations}
The structural result characterizes unchanged-ridge minimaxity and bounds
additional error from supplied shape information. Universal compatibility
differs from fixed-query optimality. Entrywise certification needs an exact
diagonal and entry bounds; unsigned local risks may not determine signs.
Full shape knowledge permits direct comparison.

Same-budget certificates bound remaining worst-case improvement, not realized
fidelity. In the retained comparison, full spectral design trades computation for tighter bounds.
Architectural constraints supply response budgets; absolute tolerances are
policies, and ellipsoid impossibility need not hold for further restricted subclasses.

Numerical results condition on cached inputs, fixed queries, supplied classes and the
arithmetic model. Reused documents add no independent corpora. Hidden
membership, backend stability, pretraining independence, population/new-query
guarantees and annotation savings remain unestablished.

\par\noindent\emph{Reproducibility.}
The English artifact retains inputs, exact matrices, sets, predictions,
protocols, executed sources and adverse outcomes. A closed manifest and
relative entry points support offline replay with declared Python/TeX
dependencies. Original provider collections and later cached-input studies retain separate
dates and source bindings.

\par\noindent\emph{Use of generative AI.}
Generative AI assisted manuscript preparation, code development, and review.
The author is responsible for the content.

\acks{The author reports no third-party funding or support for this work and
no relevant financial relationships during the preceding 36 months. The author
declares no conflicts of interest.}

\FloatBarrier
\appendix
\section{Numerical Enclosure of the Same-Budget Bounds}
\label{app:verified-arithmetic}
Stored binary64 $G=G^\top,C,\lambda$ are exact real inputs; exact
$K_*=G-\lambda I\succ0$ is verified first. Each cross-vector admits a query Gram
completion. Standard residual principles \citep{rump2010verification} certify
exact stored-matrix refitting, excluding unrounded RBF kernels and rounded predictions.

Assume binary64 round-to-nearest, gradual underflow, no overflow, and conventional
BLAS products with at most $2d$ elementary roundings per length-$d$ contributing
path. For $u=2^{-53}$, $\eta=2^{-1074}$, $2du<1$, set
$\gamma=2du/(1-2du)$ and $\zeta=2d\eta/(1-2du)$:
\[
 |\operatorname{fl}(a^\top b)-a^\top b|
 \le\gamma\sum_i|a_ib_i|+\zeta.
\]
Enclose the positive sum. Scalars round outward; norms use scaled squares/enclosed
sums. Library solves/eigenvectors are proposals; runtime probes/version records
check the contract, not arbitrary BLAS instruction streams.

For proposed Cholesky factor $L$ and inverse $T$, enclosed bounds
$\delta\ge\|I-LT\|_F$, $\delta<1$, $t\ge\|T\|_F$ and
$e\ge\|K_*-LL^\top\|_F$ give
\[
 \mu_K=(1-\delta)^2/t^2-e\le\lambda_{\min}(K_*).
\]
Positive $\mu_K$ verifies definiteness; $\mu_G=\mu_K+\lambda$ lower-bounds
$G$ and its principal blocks. For nonzero stored $w$, enclose $c=C^\top w$;
proposed solves $x,y$ give
\begin{align*}
 c^\top K_*^{-1}c
 &\le 2c^\top x-x^\top K_*x+\|c-K_*x\|_2^2/\mu_K,\\
 c^\top G^{-1}c&\ge 2c^\top y-y^\top Gy.
\end{align*}
Let $a_+$, $r_-$ and $h_+$ respectively enclose these energies above, below,
and the sum of the $k$ largest $c_i^2$ above. Monotonicity of
$ah/(\lambda a+h)$ in positive $a,h$ gives the downward lower bound
\[
 L_k=\frac{\bigl[r_- - a_+h_+/(\lambda a_++h_+)\bigr]_+}
              {(w^\top w)_+}\le\tau_k^*.
\]
Operations round in the required direction; nonpositive $L_k$ is a noncertificate.

Residual enclosure for proposed $X\approx G^{-1}C^\top$ gives
$\|CG^{-1}C^\top-CX\|_{\op}\le
\|C\|_F\|C^\top-GX\|_F/\mu_G$.
Subset solves and enclosed products/subtraction give
$\|R_S-H\|_{\op}\le\epsilon$ for exactly symmetric stored $H$.
Arbitrary proposed $Q,d$ with verified bounds
$\omega\ge\|Q^\top Q-I\|_F$ and
$e_Q\ge\|H-Q\operatorname{diag}(d)Q^\top\|_F$ imply
\[
 \tau_+=(1+\omega)\max(0,\max_j d_j)+e_Q+\epsilon
 \ge\lambda_{\max}(R_S).
\]
For $L_k>0$, evaluating $\sqrt{\tau_+/L_k}$ outward upper-bounds the radius ratio.
Exact rational checks cover cancellation, subnormals and complete constructed bounds.
Artifacts retain failed gates, code, bindings, residual allowances and all case bounds.

Proposition~\ref{prop:split-certificate} uses stored $\beta$ in the definiteness
gate, preserving exact $G,C$. Enclose $Z=C^\top F$, $K_\beta X$; any proposed
$Y$ gives reference-energy lower $2\tr(Z^\top Y)-\tr(Y^\top GY)$.
Upper-bound $\tr(X^\top K_\beta X)$ and all squared row norms of $U=Z-K_\beta X$.
Evaluate Equation~\eqref{eq:split-certificate} downward: subtract the quadratic
upper and the largest-$k$ row-bound sum divided by $\beta$ from the energy lower,
clip at zero, then divide by the upper bound on $\|F\|_F^2$.
Maximizing with a verified directional bound remains valid. The theorem
permits singular $K_\beta$; implementation requires strict positivity.
Failed positivity or nonpositive lower bounds give noncertificates, never ratios.

\section{Historical Arithmetic and Timing Conventions}
\label{app:numerical-implementation}
Section~\ref{sec:observed} uses exact-rational fallback on stored operands for
cancellation, range, definiteness, ridge and solve checks. Wider residuals
trigger refinement without interval certification; subnormal Schur differences
are normalized. Fallback bit costs exceed unit-cost counts.

Before hashing, the original single-thread run took 2390.7 seconds:
1354.6 instrumented selection, 164.8 separate validation. Greedy includes
initialization, excludes covariance checks; GLS charges shared three-order
work per preset. Phases, predictions, failures and readbacks are archived.
The 262 corrected linear-kernel GLS rows at $\alpha=10^{-2}$ leave primary rows
unchanged; corrected timings are unmeasured.

\section{Text Experiment Reproduction and Descriptive Replication}
\label{app:text-reproduction}
All 180 original geometries, 3,240 outcomes and sixteen primary slots are archived.
Tables~\ref{tab:text-small-all}--\ref{tab:text-small-task} retain all small-representation
methods; model and dimension change together. Reused documents are not independent replications.

\begin{table}[!htb]
\centering\small
\setlength{\tabcolsep}{4pt}
\begin{tabular}{@{}lrrr@{}}
\toprule
Method & BANKING77 & AG News & DBpedia14 \\
\midrule
Trace $p=1$ & $0.0446\,/\,0.0238$ & $0.0443\,/\,0.0939$ & $0.0446\,/\,0.0682$ \\
Power $p=2$ & $0.0433\,/\,0.0237$ & $0.0434\,/\,0.0956$ & $0.0434\,/\,0.0684$ \\
Power $p=4$ & $0.0426\,/\,0.0241$ & $0.0425\,/\,0.0962$ & $0.0425\,/\,0.0692$ \\
Log-det & $0.0718\,/\,0.0304$ & $0.0609\,/\,0.1107$ & $0.0548\,/\,0.0820$ \\
Ridge leverage & $0.0851\,/\,0.0361$ & $0.0690\,/\,0.1377$ & $0.0708\,/\,0.1079$ \\
Random & $0.0505\,/\,0.0276$ & $0.0483\,/\,0.1024$ & $0.0484\,/\,0.0744$ \\
\bottomrule
\end{tabular}

\caption{Small representation, $n=2048,m=512$: mean normalized radius /
query fidelity RMS across ten draws.}
\label{tab:text-small-all}
\end{table}

\begin{table}[!htb]
\centering\small
\setlength{\tabcolsep}{4pt}
\begin{tabular}{@{}lrrr@{}}
\toprule
Method & BANKING77 & AG News & DBpedia14 \\
\midrule
Full reference & $85.30\,/\,84.50$ & $90.18\,/\,90.26$ & $96.90\,/\,96.85$ \\
\addlinespace[3pt]
Trace $p=1$ & $66.08\,/\,62.09$ & $88.04\,/\,88.07$ & $92.64\,/\,92.16$ \\
Power $p=2$ & $68.20\,/\,64.54$ & $88.06\,/\,88.09$ & $91.90\,/\,91.26$ \\
Power $p=4$ & $69.14\,/\,65.88$ & $88.11\,/\,88.14$ & $93.05\,/\,92.59$ \\
Log-det & $62.64\,/\,57.01$ & $88.74\,/\,88.75$ & $94.51\,/\,94.26$ \\
Ridge leverage & $50.21\,/\,42.64$ & $85.04\,/\,84.60$ & $75.66\,/\,72.25$ \\
Random & $67.80\,/\,64.10$ & $88.49\,/\,88.54$ & $94.06\,/\,93.76$ \\
\bottomrule
\end{tabular}

\caption{Small representation: mean heldout accuracy / macro-F1 (percent).
Full KRR uses 2,048 responses; subsets use 512.}
\label{tab:text-small-task}
\end{table}

\begin{table}[!htb]
\centering\small
\setlength{\tabcolsep}{4pt}
\begin{tabular}{@{}llrr@{}}
\toprule
Data set & Power & \shortstack{Accuracy difference\\pp [95\% interval]} & \shortstack{Macro-F1 difference\\pp [95\% interval]} \\
\midrule
BANKING77 & $p=2$ & $+2.28\,[+1.69,+2.88]$ & $+2.93\,[+2.36,+3.49]$ \\
BANKING77 & $p=4$ & $+2.95\,[+2.22,+3.73]$ & $+4.28\,[+3.61,+4.95]$ \\
\addlinespace[3pt]
AG News & $p=2$ & $+0.06\,[-0.14,+0.26]$ & $+0.05\,[-0.15,+0.26]$ \\
AG News & $p=4$ & $+0.17\,[-0.10,+0.43]$ & $+0.18\,[-0.10,+0.44]$ \\
\addlinespace[3pt]
DBpedia14 & $p=2$ & $-0.33\,[-0.58,-0.10]$ & $-0.33\,[-0.60,-0.07]$ \\
DBpedia14 & $p=4$ & $+0.95\,[+0.62,+1.30]$ & $+1.06\,[+0.71,+1.42]$ \\
\bottomrule
\end{tabular}

\caption{Large-representation heldout differences to trace at
$n=2048,m=512$, with conditional 95\% paired-document percentile intervals
(percentage points). Positive values favor the stated power.}
\label{tab:text-holdout-bootstrap}
\end{table}

Table~\ref{tab:text-holdout-bootstrap} uses common document weights across fits
and full-vocabulary macro-F1. Intervals condition on fitted predictors, without
claims of source independence or adjusted primary significance. Historical input
gates used three BLAS threads; comparisons, eight single-thread workers.
Only bounds verified using Appendix~\ref{app:verified-arithmetic}'s enclosure methods
include rounding allowances.

\FloatBarrier

\FloatBarrier
\bibliography{references}

\begin{thebibliography}{31}
\providecommand{\natexlab}[1]{#1}
\providecommand{\url}[1]{\texttt{#1}}
\expandafter\ifx\csname urlstyle\endcsname\relax
  \providecommand{\doi}[1]{doi: #1}\else
  \providecommand{\doi}{doi: \begingroup \urlstyle{rm}\Url}\fi

\bibitem[Alberti et~al.(2021)Alberti, De~Vito, Lassas, Ratti, and
  Santacesaria]{alberti2021}
Giovanni~S. Alberti, Ernesto De~Vito, Matti Lassas, Luca Ratti, and Matteo
  Santacesaria.
\newblock Learning the optimal {Tikhonov} regularizer for inverse problems.
\newblock In \emph{Advances in Neural Information Processing Systems},
  volume~34, pages 25205--25216, 2021.
\newblock URL
  \url{https://proceedings.neurips.cc/paper_files/paper/2021/file/d3e6cd9f66f2c1d3840ade4161cf7406-Paper.pdf}.

\bibitem[Bach(2013)]{bach2013}
Francis Bach.
\newblock Sharp analysis of low-rank kernel matrix approximations.
\newblock In \emph{Proceedings of the 26th Annual Conference on Learning
  Theory}, volume~30 of \emph{Proceedings of Machine Learning Research}, pages
  185--209, 2013.

\bibitem[Bakhvalov(1971)]{bakhvalov1971}
Nikolai~S. Bakhvalov.
\newblock On the optimality of linear methods for operator approximation in
  convex classes of functions.
\newblock \emph{USSR Computational Mathematics and Mathematical Physics},
  11\penalty0 (4):\penalty0 244--249, 1971.
\newblock \doi{10.1016/0041-5553(71)90017-6}.

\bibitem[Besag(1974)]{besag1974}
Julian Besag.
\newblock Spatial interaction and the statistical analysis of lattice systems.
\newblock \emph{Journal of the Royal Statistical Society. Series B
  (Methodological)}, 36\penalty0 (2):\penalty0 192--225, 1974.
\newblock \doi{10.1111/j.2517-6161.1974.tb00999.x}.
\newblock URL \url{https://academic.oup.com/jrsssb/article/36/2/192/7027394}.

\bibitem[Cai et~al.(2026)Cai, Xu, and Xu]{caixuxu2026}
Jian-Feng Cai, Zhiqiang Xu, and Zili Xu.
\newblock Interlacing polynomial method for matrix approximation via
  generalized column and row selection.
\newblock \emph{Foundations of Computational Mathematics}, 26\penalty0
  (3):\penalty0 1759--1808, 2026.
\newblock \doi{10.1007/s10208-025-09719-5}.
\newblock URL \url{https://arxiv.org/abs/2312.01715v2}.
\newblock Version 2 dated 2025-04-18; first published online 2025-07-14.

\bibitem[Casanueva et~al.(2020)Casanueva, Tem{\v{c}}inas, Gerz, Henderson, and
  Vuli{\'c}]{casanueva2020banking}
I{\~n}igo Casanueva, Tadas Tem{\v{c}}inas, Daniela Gerz, Matthew Henderson, and
  Ivan Vuli{\'c}.
\newblock Efficient intent detection with dual sentence encoders.
\newblock In \emph{Proceedings of the 2nd Workshop on Natural Language
  Processing for Conversational {AI}}, pages 38--45. Association for
  Computational Linguistics, 2020.
\newblock \doi{10.18653/v1/2020.nlp4convai-1.5}.
\newblock URL \url{https://aclanthology.org/2020.nlp4convai-1.5/}.

\bibitem[Chamon and Ribeiro(2017)]{chamonribeiro2017}
Luiz F.~O. Chamon and Alejandro Ribeiro.
\newblock Approximate supermodularity bounds for experimental design.
\newblock In \emph{Advances in Neural Information Processing Systems},
  volume~30, pages 5403--5412, 2017.
\newblock URL
  \url{https://papers.nips.cc/paper_files/paper/2017/file/0d9095b0d6bbe98ea0c9c02b11b59ee3-Paper.pdf}.

\bibitem[Chen et~al.(2026)Chen, Ji, and Saibaba]{chen2026}
Jessie Chen, Hangjie Ji, and Arvind~K. Saibaba.
\newblock Optimal sensor placement in {Gaussian} processes via column subset
  selection.
\newblock arXiv:2601.20781v1, 2026.
\newblock URL \url{https://arxiv.org/abs/2601.20781v1}.
\newblock Submitted 2026-01-28.

\bibitem[Das and Kempe(2018)]{daskempe2018}
Abhimanyu Das and David Kempe.
\newblock Approximate submodularity and its applications: Subset selection,
  sparse approximation and dictionary selection.
\newblock \emph{Journal of Machine Learning Research}, 19\penalty0
  (3):\penalty0 1--34, 2018.
\newblock URL \url{https://jmlr.org/papers/v19/16-534.html}.

\bibitem[Dudley et~al.(2022)Dudley, Saibaba, and Alexanderian]{dudley2022}
Ethan Dudley, Arvind~K. Saibaba, and Alen Alexanderian.
\newblock {Monte Carlo} estimators for the {Schatten} p-norm of symmetric
  positive semidefinite matrices.
\newblock \emph{Electronic Transactions on Numerical Analysis}, 55:\penalty0
  213--241, 2022.
\newblock \doi{10.1553/etna_vol55s213}.
\newblock URL
  \url{https://etna.ricam.oeaw.ac.at/volumes/2021-2030/vol55/abstract.php?pages=213-241}.

\bibitem[El~Alaoui and Mahoney(2015)]{elalaoui2015}
Ahmed El~Alaoui and Michael~W. Mahoney.
\newblock Fast randomized kernel ridge regression with statistical guarantees.
\newblock In \emph{Advances in Neural Information Processing Systems},
  volume~28, pages 775--783, 2015.

\bibitem[Elenberg et~al.(2018)Elenberg, Khanna, Dimakis, and
  Negahban]{elenberg2018}
Ethan~R. Elenberg, Rajiv Khanna, Alexandros~G. Dimakis, and Sahand Negahban.
\newblock Restricted strong convexity implies weak submodularity.
\newblock \emph{Annals of Statistics}, 46\penalty0 (6B):\penalty0 3539--3568,
  2018.
\newblock \doi{10.1214/17-AOS1679}.
\newblock URL \url{https://eelenberg.github.io/weak-submodular-preprint.pdf}.

\bibitem[Farahat et~al.(2013)Farahat, Ghodsi, and Kamel]{farahat2013}
Ahmed~K. Farahat, Ali Ghodsi, and Mohamed~S. Kamel.
\newblock A fast greedy algorithm for generalized column subset selection.
\newblock arXiv:1312.6820v1, 2013.
\newblock URL \url{https://arxiv.org/abs/1312.6820v1}.
\newblock NIPS'13 Workshop on Greedy Algorithms, Frank-Wolfe and Friends.

\bibitem[Foucart and Liao(2023)]{foucartliao2023}
Simon Foucart and Chunyang Liao.
\newblock Optimal recovery from inaccurate data in {Hilbert} spaces:
  Regularize, but what of the parameter?
\newblock \emph{Constructive Approximation}, 57\penalty0 (2):\penalty0
  489--520, 2023.
\newblock \doi{10.1007/s00365-022-09590-5}.
\newblock URL
  \url{https://link.springer.com/article/10.1007/s00365-022-09590-5}.

\bibitem[Foucart et~al.(2022)Foucart, Liao, Shahrampour, and Wang]{foucart2022}
Simon Foucart, Chunyang Liao, Shahin Shahrampour, and Yinsong Wang.
\newblock Learning from non-random data in {Hilbert} spaces: an optimal
  recovery perspective.
\newblock \emph{Sampling Theory, Signal Processing, and Data Analysis}, 20,
  2022.
\newblock \doi{10.1007/s43670-022-00022-w}.
\newblock Article 5.

\bibitem[Foucart et~al.(2023)Foucart, Liao, and Veldt]{foucartliaoveldt2023}
Simon Foucart, Chunyang Liao, and Nate Veldt.
\newblock On the optimal recovery of graph signals.
\newblock In \emph{2023 International Conference on Sampling Theory and
  Applications (SampTA)}, pages 1--5, 2023.
\newblock \doi{10.1109/SampTA59647.2023.10301205}.
\newblock URL \url{https://ieeexplore.ieee.org/document/10301205}.

\bibitem[Gama et~al.(2020)Gama, Marques, Mateos, and Ribeiro]{gama2020}
Fernando Gama, Antonio~G. Marques, Gonzalo Mateos, and Alejandro Ribeiro.
\newblock Rethinking sketching as sampling: A graph signal processing approach.
\newblock \emph{Signal Processing}, 169:\penalty0 107404, 2020.
\newblock \doi{10.1016/j.sigpro.2019.107404}.

\bibitem[Jaggi(2013)]{jaggi2013}
Martin Jaggi.
\newblock Revisiting {Frank--Wolfe}: Projection-free sparse convex
  optimization.
\newblock In \emph{Proceedings of the 30th International Conference on Machine
  Learning}, volume~28 of \emph{Proceedings of Machine Learning Research},
  pages 427--435. PMLR, 2013.
\newblock URL \url{https://proceedings.mlr.press/v28/jaggi13.html}.

\bibitem[Joshi and Boyd(2009)]{joshiboyd2009}
Siddharth Joshi and Stephen Boyd.
\newblock Sensor selection via convex optimization.
\newblock \emph{IEEE Transactions on Signal Processing}, 57\penalty0
  (2):\penalty0 451--462, 2009.
\newblock \doi{10.1109/TSP.2008.2007095}.
\newblock URL
  \url{https://web.stanford.edu/~boyd/papers/sensor_selection.html}.

\bibitem[Kanagawa et~al.(2018)Kanagawa, Hennig, Sejdinovic, and
  Sriperumbudur]{kanagawa2018}
Motonobu Kanagawa, Philipp Hennig, Dino Sejdinovic, and Bharath~K.
  Sriperumbudur.
\newblock {Gaussian} processes and kernel methods: A review on connections and
  equivalences, 2018.
\newblock arXiv:1807.02582v1.

\bibitem[Krause et~al.(2008)Krause, Singh, and Guestrin]{krause2008}
Andreas Krause, Ajit Singh, and Carlos Guestrin.
\newblock Near-optimal sensor placements in {Gaussian} processes: Theory,
  efficient algorithms and empirical studies.
\newblock \emph{Journal of Machine Learning Research}, 9\penalty0 (8):\penalty0
  235--284, 2008.

\bibitem[Maas et~al.(2011)Maas, Daly, Pham, Huang, Ng, and Potts]{maas2011imdb}
Andrew~L. Maas, Raymond~E. Daly, Peter~T. Pham, Dan Huang, Andrew~Y. Ng, and
  Christopher Potts.
\newblock Learning word vectors for sentiment analysis.
\newblock In \emph{Proceedings of the 49th Annual Meeting of the Association
  for Computational Linguistics: Human Language Technologies}, pages 142--150.
  Association for Computational Linguistics, 2011.
\newblock URL \url{https://aclanthology.org/P11-1015/}.

\bibitem[Micchelli and Rivlin(1977)]{micchelli1977}
Charles~A. Micchelli and Theodore~J. Rivlin.
\newblock A survey of optimal recovery.
\newblock In \emph{Optimal Estimation in Approximation Theory}, pages 1--54.
  Springer, 1977.
\newblock \doi{10.1007/978-1-4684-2388-4_1}.

\bibitem[Newman(1960)]{newman1960}
Morris Newman.
\newblock {Kantorovich}'s inequality.
\newblock \emph{Journal of Research of the National Bureau of Standards,
  Section B: Mathematics and Mathematical Physics}, 64B\penalty0 (1):\penalty0
  33--34, 1960.
\newblock \doi{10.6028/jres.064B.003}.

\bibitem[{OpenAI}(2024)]{openai2024embeddings}
{OpenAI}.
\newblock New embedding models and {API} updates, January 2024.
\newblock URL
  \url{https://openai.com/index/new-embedding-models-and-api-updates/}.
\newblock Released January 25, 2024.

\bibitem[Ordozgoiti et~al.(2022)Ordozgoiti, Matakos, and
  Gionis]{ordozgoiti2022}
Bruno Ordozgoiti, Antonis Matakos, and Aristides Gionis.
\newblock Generalized leverage scores: Geometric interpretation and
  applications.
\newblock In \emph{Proceedings of the 39th International Conference on Machine
  Learning}, volume 162 of \emph{Proceedings of Machine Learning Research},
  pages 17056--17070, 2022.
\newblock URL \url{https://proceedings.mlr.press/v162/ordozgoiti22a.html}.

\bibitem[Rump(2010)]{rump2010verification}
Siegfried~M. Rump.
\newblock Verification methods: Rigorous results using floating-point
  arithmetic.
\newblock \emph{Acta Numerica}, 19:\penalty0 287--449, 2010.
\newblock \doi{10.1017/S096249291000005X}.

\bibitem[Sagnol(2013)]{sagnol2013}
Guillaume Sagnol.
\newblock Approximation of a maximum-submodular-coverage problem involving
  spectral functions, with application to experimental designs.
\newblock \emph{Discrete Applied Mathematics}, 161\penalty0 (1--2):\penalty0
  258--276, 2013.
\newblock \doi{10.1016/j.dam.2012.07.016}.
\newblock URL \url{https://www.zib.de/userpage/sagnol/publication/sag13_dam/}.

\bibitem[{scikit-learn developers}(n.d.)]{sklearn_toy180}
{scikit-learn developers}.
\newblock \emph{Toy datasets}, n.d.
\newblock URL \url{https://scikit-learn.org/1.8/datasets/toy_dataset.html}.
\newblock scikit-learn 1.8.0 documentation.

\bibitem[Ver~Hoef et~al.(2018)Ver~Hoef, Hanks, and Hooten]{verhoef2018}
Jay~M. Ver~Hoef, Ephraim~M. Hanks, and Mevin~B. Hooten.
\newblock On the relationship between conditional ({CAR}) and simultaneous
  ({SAR}) autoregressive models.
\newblock \emph{Spatial Statistics}, 25:\penalty0 68--85, 2018.
\newblock \doi{10.1016/j.spasta.2018.04.006}.
\newblock URL \url{https://arxiv.org/abs/1710.07000v1}.

\bibitem[Zhang et~al.(2015)Zhang, Zhao, and LeCun]{zhang2015text}
Xiang Zhang, Junbo Zhao, and Yann LeCun.
\newblock Character-level convolutional networks for text classification.
\newblock In \emph{Advances in Neural Information Processing Systems},
  volume~28, pages 649--657, 2015.
\newblock URL
  \url{https://papers.nips.cc/paper/2015/hash/250cf8b51c773f3f8dc8b4be867a9a02-Abstract.html}.

\end{thebibliography}
\end{document}